\documentclass[journal]{IEEEtran}
\ifCLASSINFOpdf
\else
\fi

\usepackage[utf8]{inputenc}
\usepackage[T1]{fontenc}
\usepackage[pdftex]{graphicx}
\usepackage{graphicx}
\usepackage{tikz}
\usepackage{subcaption}
\graphicspath{{../Figures/}}
\usepackage{multirow}
\usepackage{float}
\usepackage{graphics} 
\usepackage{multicol}
\usepackage{subcaption}
\usepackage{color, colortbl}
\usepackage[]{algorithm2e}
\usepackage{amsmath} 
\usepackage{array}  
\usepackage{threeparttable}
\usepackage{bm} 

\begin{document}
%
\title{Camera-Lidar Integration: Probabilistic sensor fusion for semantic mapping}
%


%
%

\author{Julie Stephany Berrio,~\IEEEmembership{Member,~IEEE,} 
        Mao Shan,~\IEEEmembership{Member,~IEEE,} 
        \\
        Stewart Worrall,~\IEEEmembership{Member,~IEEE,}
        and Eduardo Nebot,~\IEEEmembership{Fellow,~IEEE}

\thanks{J. Berrio, M. Shan, S. Worrall and E. Nebot are with the 
Australian Centre for Field Robotics (ACFR) at the University of Sydney, NSW, Australia. E-mails: {\tt \{ j.berrio, m.shan, s.worrall, e.nebot\}@acfr.USyd.edu.au}.}
\thanks{Manuscript submitted in March, 2020.}}

%
%

\markboth{Submission for XXX Journal}%
{Shell \MakeLowercase{\textit{et al.}}: Bare Demo of IEEEtran.cls for IEEE Journals}
%



\maketitle

\begin{abstract}
An automated vehicle operating in an urban environment must be able to perceive and recognise object/obstacles in a three-dimensional world while navigating in a constantly changing environment. 
In order to plan and execute accurate sophisticated driving maneuvers a high-level contextual understanding of the surroundings is essential. 
Due to the recent progress in image processing, it is now possible to obtain high definition semantic information in 2D from monocular cameras, though cameras cannot reliably provide the highly accurate 3D information provided by lasers.  
The fusion of these two sensor modalities can overcome the shortcomings of each individual sensor, though there are a number of important challenges that need to be addressed in a probabilistic manner.
In this paper we address the common, yet challenging, lidar/camera/semantic fusion problems which are seldom approached in a wholly probabilistic manner.
Our approach is capable of using a multi-sensor platform to build a three-dimensional semantic voxelized map that considers the uncertainty of all of the processes involved.
We present a probabilistic pipeline that incorporates uncertainties from the sensor readings (cameras, lidar, IMU and wheel encoders), compensation for the motion of the vehicle, and heuristic label probabilities for the semantic images.
We also present a novel and efficient viewpoint validation algorithm to check for occlusions from the camera frames. 
A probabilistic projection is performed from the camera images to the lidar point cloud. 
Each labelled lidar scan then feeds into an octree map building algorithm that updates the class probabilities of the map voxels every time a new observation is available. 
We validate our approach using a set of qualitative and quantitative experimental tests on the USyd Dataset \cite{usyd_dataset}. 
These tests demonstrate the usefulness of a probabilistic sensor fusion approach by evaluating the performance of the perception system in a typical autonomous vehicle application.
\end{abstract}


%
\IEEEpeerreviewmaketitle

\section{Introduction}

\IEEEPARstart{B}{uilding} a sufficiently descriptive map of the environment is a fundamental challenge for any autonomous system operating in complex and dynamic environments. 
To enable successful self-driving car applications, comprehensive information (such as location and type of the objects) is required for creating high level maps of the surrounding environment to ensure both accuracy/reliability in localization, and also safe and optimal path planning and navigation. 
The semantic meaning of the objects surrounding the vehicle is one of the most significant pieces of information necessary for the decision making process. 

\begin{figure}[!h]
\vspace{3mm}
\centerline{
\includegraphics[width=0.99\columnwidth]{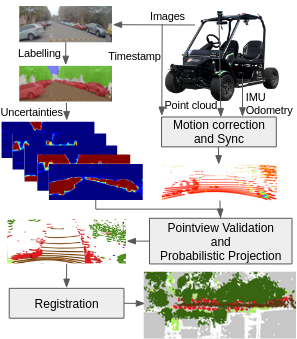}
}
\caption{\small Pipeline of the proposed process. Images and a point cloud measurements of the environment are collected by an electric vehicle retrofitted with a comprehensive autonomous sensor suite. The images are semantically segmented while the point cloud is corrected for motion. Uncertainty information is considered in the projection of the label information into the corrected point cloud. An octree map building algorithm is used to register the point cloud. }
\label{fig:pipeline_process}
\end{figure}


Cameras are sensors that are commonly used for building a high level understanding of the urban environment. 
Cameras are affordable, have low power consumption and contain texture and colour data, which make them an effective sensor to be used for object classification \cite{ramos_2015}. 

During the last few years, there has been significant progress in semantic vision. It is now possible to get accurate 2D semantic segmentation in real time. In this paper, we use vision information as input for a trained CNN \cite{wie_2019} which outputs images incorporating the semantic labels of objects in the image. We have also extended our previous work \cite{stephany_iros} and developed a heuristic method to associate uncertainty to each pixel class, by modifying the CNN's final softmax layer based on the label distribution within the raw image's superpixels. 

Nevertheless, when using only cameras it is difficult to obtain accurate 3D position information of the detected objects, particularly at longer range. 
To overcome this problem, lidar sensors have been widely used in autonomous vehicles research to provide robust, high quality 3D information. 
This is particularly noticeable at longer range, with lidars able to accurately determine the range of objects at greater than 100 metres.
Unfortunately, currently available lidar have a much lower resolution compared to cameras, and are unable to provide color or texture information. These characteristics are important for computing relevant object descriptors/features for classification. 

By combining lidar and camera information it is possible to get a comprehensive understanding of the environment. Through camera-lidar sensor fusion, we are able to transfer relevant data from camera to lidar and vice versa, allowing a better understanding of the surrounding scene structure \cite{Hsiang_2016}.
In an ideal world, the cameras would have perfect calibration; lidars would be capable of providing a 3D point cloud taken in a single timestamp synchronised with the camera images; cameras and lidars would be located in the same reference frame, having exactly the same viewpoint; and the sensors readings and data classification would be without error.

\begin{figure}[!t]
\centerline{
\includegraphics[width=0.85\columnwidth]{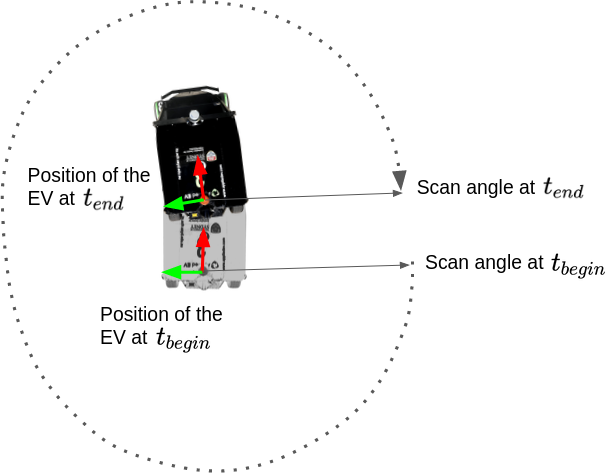}
}
\caption{\small The lidar scan traced over one revolution while the experimental platform is moving. The projection is a circle when the vehicle is stationary, and a more complex shape when the vehicle is moving.}
\label{fig:motion_scan}
\end{figure}

\begin{figure}[!t]
\centerline{
\includegraphics[width=\columnwidth]{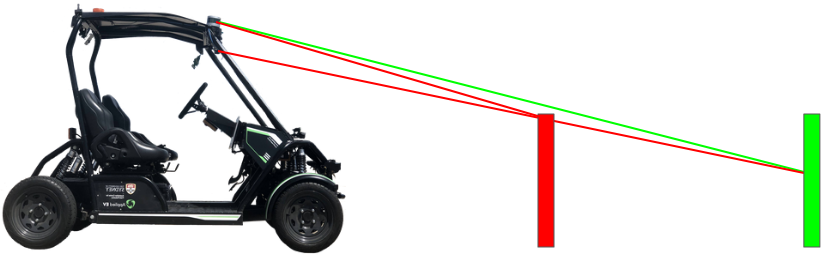}
}
\caption{\small Given that the lidar is located above the camera, both sensors have a slightly different point of view. Due to this, the red and green rays project onto the same image pixel, establishing that both 3D points belong to the red object. In this case, from the camera perspective, the green object is occluded.}
\label{fig:occlusion_problem}
\end{figure}

In reality, sensors have a limited field of view and are located in different places around the vehicle.
There are also challenges when using scanning sensors, such as most lidars, as the information is not obtained instantaneously. 
To accurately fuse the information from the lidar and cameras, it is essential to address the following problems:

\begin{enumerate}
  \item \textbf{Accurate parameters}: The calibration process is important in determining the intrinsic camera transform between the real world and pixel coordinate systems, and the extrinsic lidar-camera geometric relationships between sensors.
  
  \item \textbf{Sensor synchronization and motion correction}: The multi-beam lidar scans the environment in 3D while the vehicle is moving and capturing images.
  Due to the translation and rotation of the vehicle during a single revolution of the lidar scanner, the final section of the point cloud is registered to a different location compared with the beginning, as shown in Fig. \ref{fig:motion_scan}.
  Each part of the point cloud represents the lidar returns from the perspective of the vehicle at a slightly different location.
  The camera images are an instantaneous snapshot of the environment with a timestamp that is not aligned to the lidar timestamp.
  The lidar points therefore must be compensated for motion of the vehicle and transformed into a common reference frame with the camera images before sensor fusion can take place.
  
  \item \textbf{Occlusion handling}: The occlusion problem arises because the lidar and camera can not be in the same physical location (differing perspectives), or because the lidar is capable of getting secondary returns from beyond the first detected object.
  The consequence of this means that a lidar situated higher than the camera will be able to see behind obstacles that occlude the view of the camera (being mounted lower). This leads to an incorrect transference of camera-lidar information due to incorrectly assigning the 3D points which are not visible to the camera lens. This issue is more evident in cameras mounted further from the lidar as both sensors would have an increasingly different perspective of the surrounding environment. Fig. \ref{fig:occlusion_problem} shows an example where the occlusion problem is evident. It is essential to understand this problem when selecting the location of the sensor since it restricts the common field of view in a manner that is not immediately apparent.
  
  \item \textbf{Incorporating uncertainty into the pipeline}: In a process where several pieces of sensor information are fused, it is essential to propagate their uncertainties throughout the processing pipeline.
  Estimating the uncertainty of the components in the pipeline enables the evaluation and quantification of risk. This information is crucial in a subsequent risk assessment that can be used for decision making.
\end{enumerate}
  
This paper proposes a comprehensive and novel probabilistic pipeline for the fusion of semantically labelled images and lidar point clouds. This pipeline also addresses the motion distortion and the occlusion problem. Most importantly, this approach integrates the uncertainties of all sources of information from the different processes involved. The corrected point clouds are then used to create a 3D semantic map using a modified octree framework \cite{octomap}. 
We validate our approach using the USyd Dataset \cite{dataset_paper}, which was collected by an electric vehicle shown in Fig.\ref{fig:platform}. Our main contributions are:

\begin{figure}[!t]
\vspace{3mm}
\centerline{
\includegraphics[width=0.9\columnwidth]{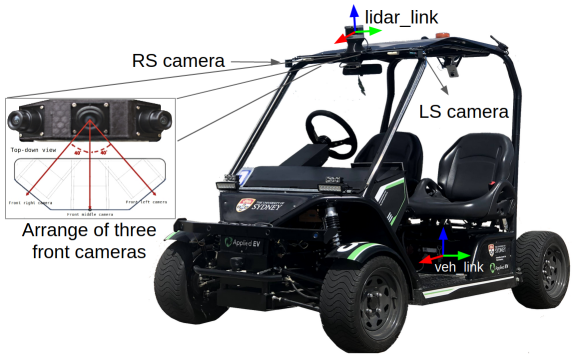}
}
\caption{\small Experimental platform equipped with five cameras (two side cameras and an arrangement of three cameras facing the front) running at 30 Hz, one Velodyne VLP-16 lidar (16 laser beams) with a frequency of 10 Hz, an IMU and wheel encoders at a rate of 100 Hz. }
\label{fig:platform}
\end{figure}

\begin{itemize}
  \item Heuristic uncertainty association to labelled semantic images.
  \item Motion correction of individual lidar packets with uncertainty. 
  \item Occlusion handling for sparse lidar point clouds required for the camera-lidar projection process. 
  \item Probabilistic pipeline that goes from camera-lidar data to a 3D map. 
\end{itemize}

The paper is organized as follows: In the next section, we present background work related to lidar/camera sensor fusion. Section III goes into detail of each component from the proposed pipeline. This includes subsections for determining the heuristic label probability, calibration, motion correction, image to lidar projection, and the registering of the point cloud to a map.
Experiments results are presented in Section IV providing a qualitative and quantitative evaluation of the algorithms. 
Finally, the conclusion and future work are presented in Section V.

\section{Related Work}

In this section, we present previous work that addresses some of the fundamental challenges of fusing camera and lidar information. Most of these contributions address part of this challenge, but very few evaluate and propagate a measure of the uncertainty within these processes.

Li et. al \cite{Li_2017} presents an analysis of the difficulty in pixel based segmentation for deep models. In this work, the pixels were divided into three categories depending on their difficulty to be segmented. Those pixels within the \textit{extremely hard} category which were mostly located on the objects' boundaries, were the main source of misclassification. Boundary pixels are very hard to classify accurately due to their large ambiguity.   

Various papers have addressed the sensor fusion between cameras and lidars using the geometric relationship between a pin-hole camera and the lidar.

Xuan et al. \cite{2007_Xuan} presents a combined (rigidly mounted) sensor system consisting of a lidar and camera for an ortho-photo mapping application. In this paper, the authors also make use of GPS and IMU observations to improve the absolute position of each image.
In \cite{2015_li}, the authors present an indoor scene construction technique which uses a 2D lidar and a digital camera.  Both sensors were mounted rigidly on a frame, and the sensor fusion is performed by using the extrinsic calibration parameters. 
In \cite{Song_2016}, authors propose a relative-localization approach using an RGB-D camera and lidar sensor, where the RGB-D and lidar measurements are fused using an extrinsic calibration algorithm.

Authors in \cite{Kang_2017} detect potholes using two 2D lidars and a single camera, the point cloud and image information is used to obtain the width and depth of the pothole. In \cite{Zhong_2017} authors perform the transference of semantic labels from an image to a point cloud using a Gauss Kernel, which takes into consideration adjacent pixels of the labeled image.
Shimizu et al. in \cite{Shimizu_2018} described a person-searching algorithm using an omnidirectional camera and one lidar. The image information is used to detect the person while signboards are identified by the lidar.  
Later, Bybee et al., in \cite{Bybee_2019} presented a bundle adjustment technique to fuse information from a low-cost lidar and camera to create terrain models. A multi-object tracking technique which rigidly fuses object proposals across sensors is presented in \cite{Rangesh_2019}.

Schneider et al. in \cite{Schneider_2010} addressed the synchronization and motion correction problem by triggering the cameras such that the image is captured when the laser-beams register in the camera field of view, and translating the scan points based on the vehicle movement. In \cite{Bayerl_2014}, authors present a method to detect and track rural crossroads combining vision and lidar measurements. In addition to synchronizing both sensors, the point cloud is concatenated to obtain a denser point cloud.  
Rieken et al. in \cite{Rieken_2016}, presented an approach to incorporate the ego-motion into a grid-based environment representation. The authors propose an adaptive prediction horizon for the object tracking algorithms, based on sensor scan timing characteristics. Authors in \cite{Sommer_2017}, presented a hardware-system for calibrating the time offsets between actual measurements of cameras and lidar, incorporating their corresponding timestamps.

In \cite{8317846_supersensor}, the synchronisation timestamp is chosen to coincide with the timestamp of the most recent camera frame, then transforming each point using the ego-motion transform matrix of the vehicle. 
Le Gentil et al. in \cite{Gentil_2018} introduce a framework to obtain the extrinsic calibration parameters of a lidar-IMU sensor system. The motion distortion in each lidar scan is also characterised in this process. In \cite{Tang_2018}, authors demonstrated the learning of a bias correction for the lidar motion estimate based on a Gaussian process. 
Authors in \cite{Hu_2018} \cite{Kokovkina_2019}, make use of the ROS \cite{ros} message filter to match different information sources up to some epsilon time difference.

The occlusion problem in camera-lidar applications becomes important when projecting information from one sensor frame to the other. The most common solution to the occlusion problem is based on segmenting the point cloud and then computing a 2D convex hull for every cluster in the image frame. This process is followed by an occlusion check, which takes into account the depth of each cluster to select which part of the point cloud is not occluded to the camera view \cite{Schneider_2010}. In \cite{Pintus_2011}, the authors presented an approach where a cone is fitted to each 3D point along the projection line with the condition that it does not intersect any other point. The aperture of the cone has to be larger than a threshold to set the point as visible.
These techniques are designed for a dense point cloud which allows the segmentation algorithm to provide consistent results. Nevertheless, for sparse point cloud as the one provided by the VLP-16, only very close objects to the vehicle can be reliably segmented, considering that these techniques not suitable in urban-type environments.

In \cite{2016_deil}, an EKF (extended Kalman filter) is used to fuse information from a 2D lidar with a camera and IMU (inertial measurement unit) for pose estimation. 
A probabilistic method for fusing 3D lidar data with stereo images was introduced in \cite{Maddern_2016}.
Dieterle et al. in \cite{Dieterle_2017} introduced camera/lidar fusion for pedestrian detection using a hierarchical data fusion approach.
Authors in \cite{Charika_2019} presented a method to predict the uncertainty of a moving lidar point cloud projected into an image. 

We referred to various publications presenting approaches capable of addressing some of the well-known problems present in lidar-camera fusion. To the best of the author's knowledge, there has not been an attempt to formulate a comprehensive approach that considers uncertainties in all processes involved with both sensing modalities.  In this work, we demonstrate how we can incorporate uncertainties from both vision-based semantic and lidar motion correction to generate a rich and consistent 3D representation of the environment.

\section{Methodology}
The pipeline showing the process is presented in Figure \ref{fig:pipeline_process}. The uncertainties corresponding to image labeling and motion correction are used to obtain a probabilistic projection.
An ENet CNN which has been fine-tuned with annotated local images and by applying data augmentation techniques \cite{wie_2019} is adopted as semantic classifier. The output of the CNN and its internal information is used to calculate the heuristic uncertainty associated with each label. We used a calibration process to determine the intrinsic parameters of the cameras, and the transformation matrix between the cameras and the lidar. The accuracy of these constants are critical for the data fusion process. 

The complete lidar scan is translated based on the motion of the vehicle and the camera timestamp. We then calculate the corresponding image coordinate for each lidar return. This process makes use of an Unscented Transform (UT) to propagate the uncertainties from the lidar sensor to the pixel coordinates in each camera image. 

\begin{figure*}[h!]
\vspace{3mm}
\centering

\begin{subfigure}[]{0.48\columnwidth}
\centering
	\includegraphics[width=\columnwidth, height=2.4cm]{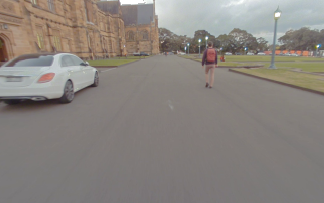}
    \caption{Original image}
    \label{sub_a}
    \end{subfigure}
\begin{subfigure}[]{0.48\columnwidth}
\centering
	\includegraphics[width=\columnwidth, height=2.4cm]{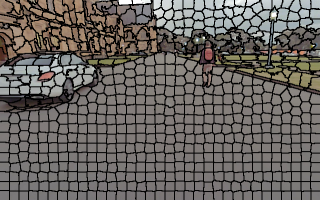}
    \caption{Superpixel clustering}
    \label{sub_b}
    \end{subfigure}
\begin{subfigure}[]{0.48\columnwidth}
\centering
	\includegraphics[width=\columnwidth, height=2.4cm]{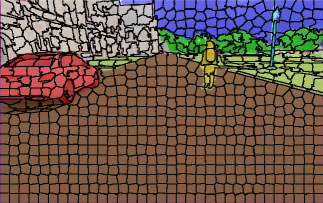}
    \caption{Semantic labels and Sp}
    \label{sub_c}
    \end{subfigure}
\begin{subfigure}[]{0.48\columnwidth}
\centering
	\includegraphics[width=\columnwidth, height=2.5cm]{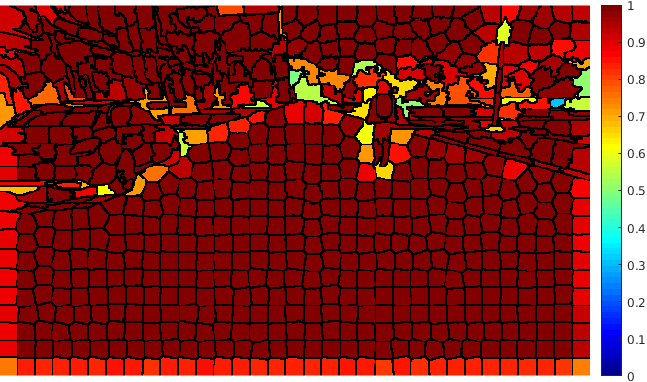}
    \caption{$\%$ of the label mode per Sp}
    \label{sub_d}
    \end{subfigure}

\caption{\small Uncertainty association process. The raw image captured by the camera is shown in \ref{sub_a}. \ref{sub_b} shows the clustering performed by the SLIC algorithm to the original image. In \ref{sub_c} the CNN's semantic segmentation result is overlaid with the superpixel segmentation. Here the colour code is: red for \textit{vehicle}, white for \textit{building}, brown for \textit{road}, green for \textit{vegetation}, blue for \textit{sky}, lime is for \textit{undrivable road}, yellow for \textit{pedestrian} and \textit{riders}, cyan for \textit{pole}, gray for \textit{fence} and purple for \textit{unlabeled pixels}, for a total of 12 classes. \ref{sub_d} displays the result of $spp_k$ within the superpixels, the color bar represents percentage of the most common label within the superpixel.}
\label{fig:proceso}
\end{figure*}

\begin{figure*}[t!]
\vspace{3mm}
\centering

\begin{subfigure}[]{0.32\columnwidth}
\centering
	\includegraphics[trim={0 0 2cm 0},clip,width=\columnwidth]{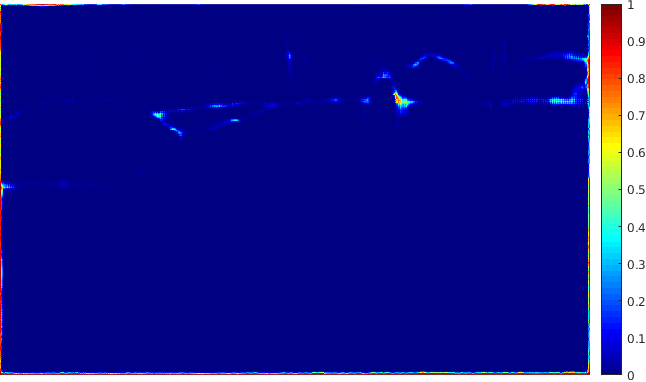}
    \caption{Unlabled}
    \label{sub_a1}
    \end{subfigure}
\begin{subfigure}[]{0.32\columnwidth}
\centering
	\includegraphics[trim={0 0 2cm 0},clip,width=\columnwidth]{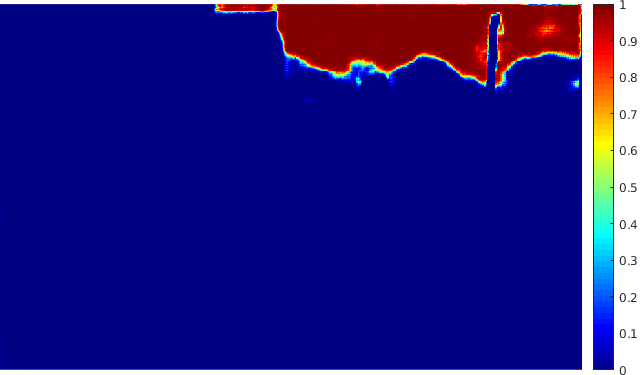}
    \caption{Sky}
    \label{sub_b1}
    \end{subfigure}
\begin{subfigure}[]{0.32\columnwidth}
\centering
	\includegraphics[trim={0 0 2cm 0},clip,width=\columnwidth]{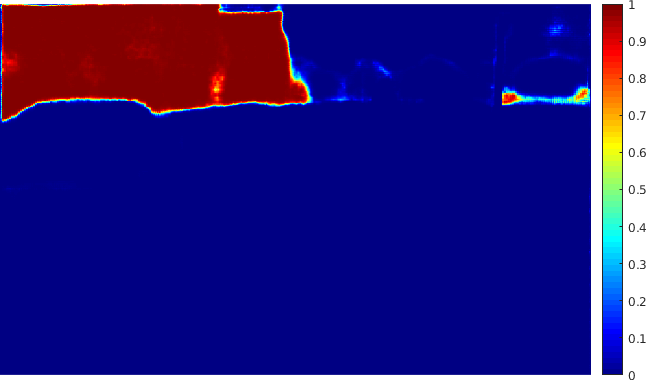}
    \caption{Building}
    \label{sub_c1}
    \end{subfigure}
\begin{subfigure}[]{0.32\columnwidth}
\centering
	\includegraphics[trim={0 0 2cm 0},clip,width=\columnwidth]{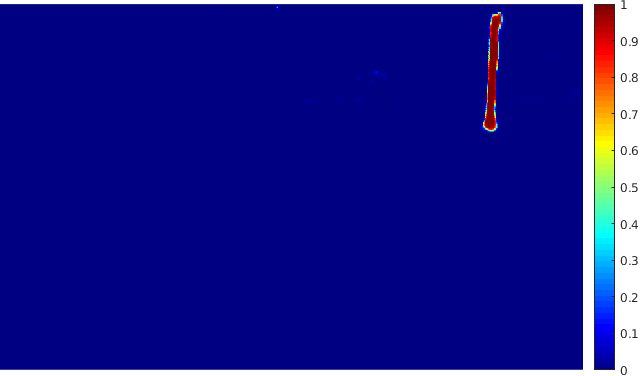}
    \caption{Pole}
    \label{sub_d1}
    \end{subfigure}
\begin{subfigure}[]{0.32\columnwidth}
\centering
	\includegraphics[trim={0 0 2cm 0},clip,width=\columnwidth]{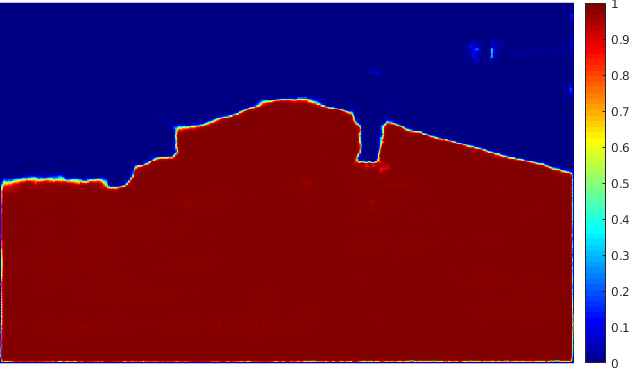}
    \caption{Road}
    \label{sub_e1}
    \end{subfigure}
\begin{subfigure}[]{0.32\columnwidth}
\centering
	\includegraphics[trim={0 0 2cm 0},clip,width=\columnwidth]{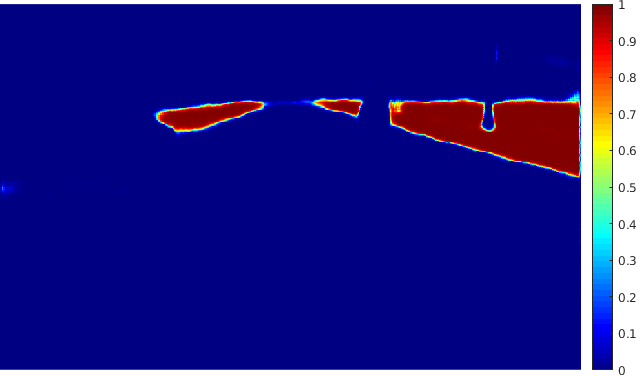}
    \caption{U. Road}
    \label{sub_f1}
    \end{subfigure}

\begin{subfigure}[]{0.32\columnwidth}
\centering
	\includegraphics[trim={0 0 2cm 0},clip,width=\columnwidth]{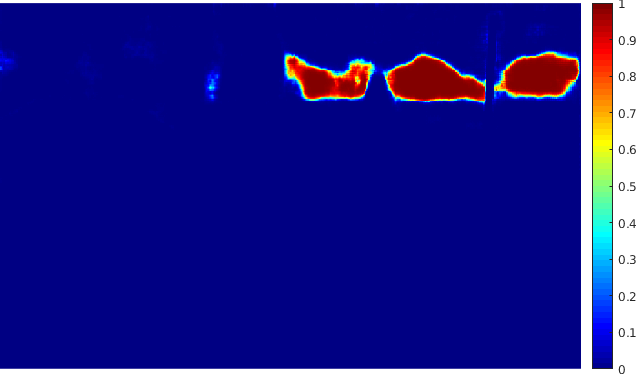}
    \caption{Vegetation}
    \label{sub_g1}
    \end{subfigure}
\begin{subfigure}[]{0.32\columnwidth}
\centering
	\includegraphics[trim={0 0 2cm 0},clip,width=\columnwidth]{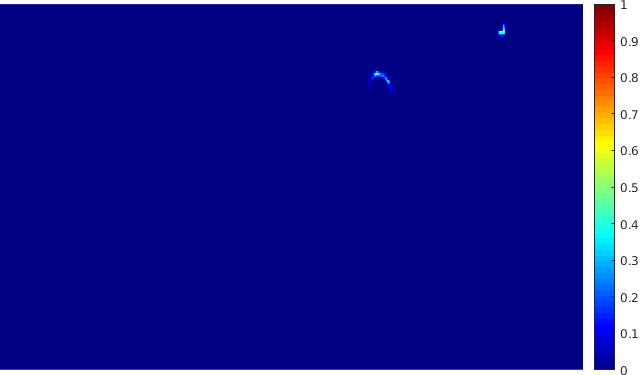}
    \caption{Sign}
    \label{sub_h1}
    \end{subfigure}
\begin{subfigure}[]{0.32\columnwidth}
\centering
	\includegraphics[trim={0 0 2cm 0},clip,width=\columnwidth]{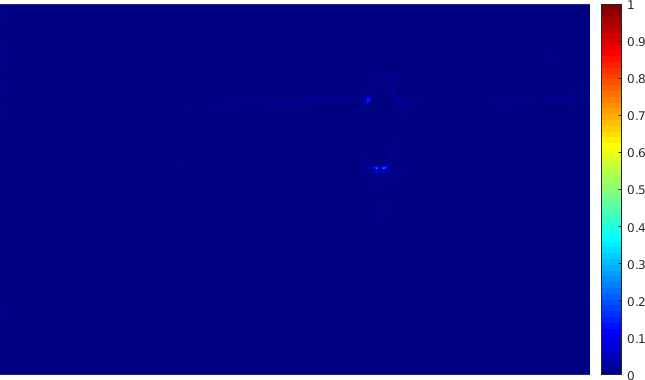}
    \caption{Fence}
    \label{sub_i1}
    \end{subfigure}
\begin{subfigure}[]{0.32\columnwidth}
\centering
	\includegraphics[trim={0 0 2cm 0},clip,width=\columnwidth]{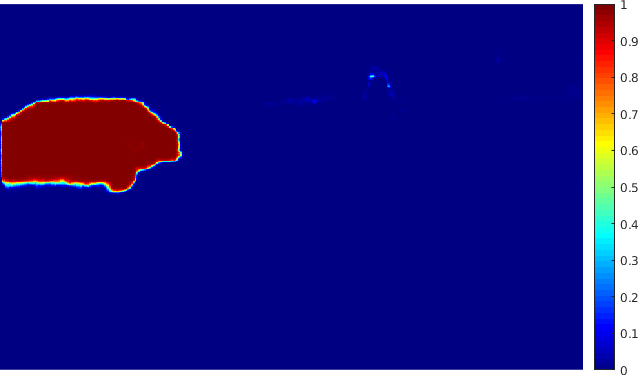}
    \caption{Vehicle}
    \label{sub_j1}
    \end{subfigure}
\begin{subfigure}[]{0.32\columnwidth}
\centering
	\includegraphics[trim={0 0 2cm 0},clip,width=\columnwidth]{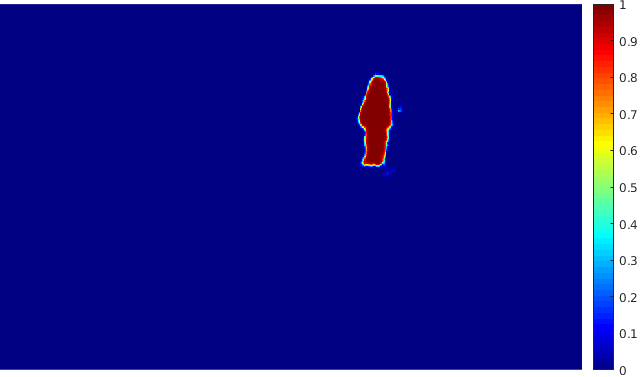}
    \caption{Pedestrian}
    \label{sub_k1}
    \end{subfigure}
\begin{subfigure}[]{0.32\columnwidth}
\centering
	\includegraphics[trim={0 0 2cm 0},clip,width=\columnwidth]{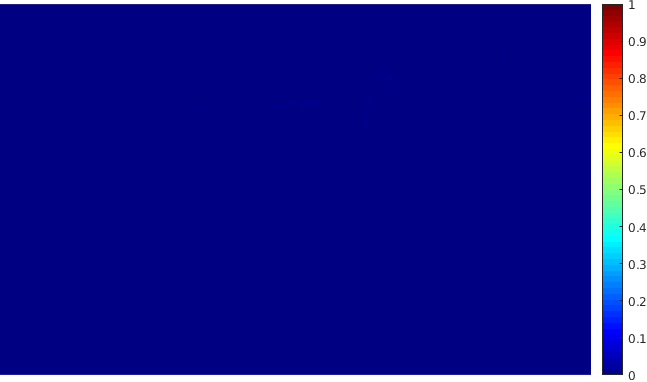}
    \caption{Rider}
    \label{sub_l1}
    \end{subfigure}

\begin{subfigure}[]{0.32\columnwidth}
\centering
	\includegraphics[trim={0 0 2cm 0},clip,width=\columnwidth]{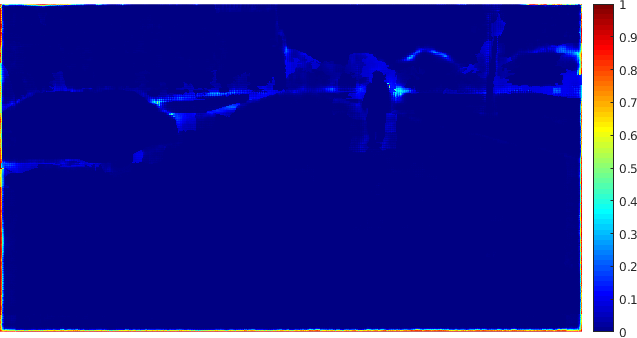}
    \caption{Unlabeled}
    \label{sub_m1}
    \end{subfigure}
\begin{subfigure}[]{0.32\columnwidth}
\centering
	\includegraphics[trim={0 0 2cm 0},clip,width=\columnwidth]{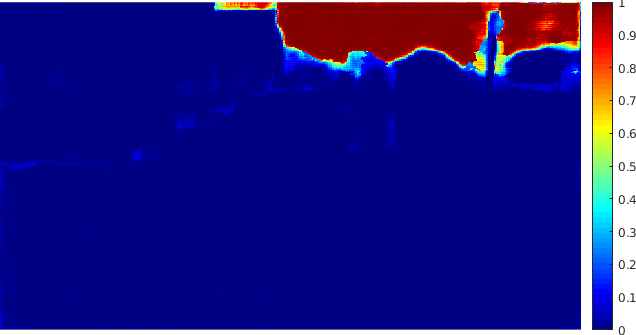}
    \caption{Sky}
    \label{sub_n1}
    \end{subfigure}
\begin{subfigure}[]{0.32\columnwidth}
\centering
	\includegraphics[trim={0 0 2cm 0},clip,width=\columnwidth]{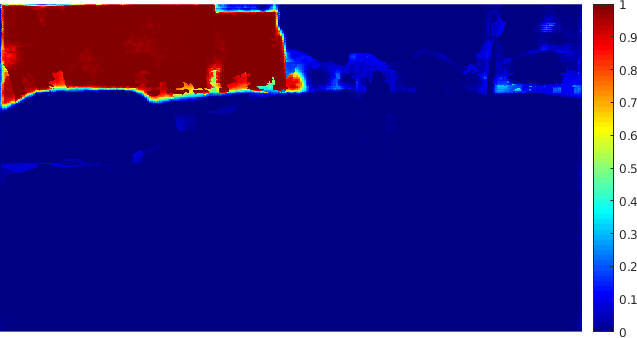}
    \caption{Building}
    \label{sub_o1}
    \end{subfigure}
\begin{subfigure}[]{0.32\columnwidth}
\centering
	\includegraphics[trim={0 0 2cm 0},clip,width=\columnwidth]{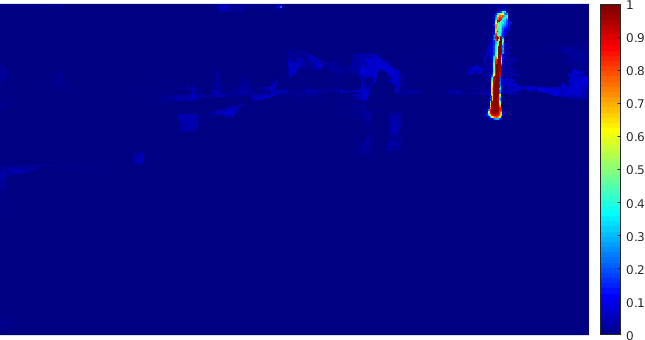}
    \caption{Pole}
    \label{sub_p1}
    \end{subfigure}
\begin{subfigure}[]{0.32\columnwidth}
\centering
	\includegraphics[trim={0 0 2cm 0},clip,width=\columnwidth]{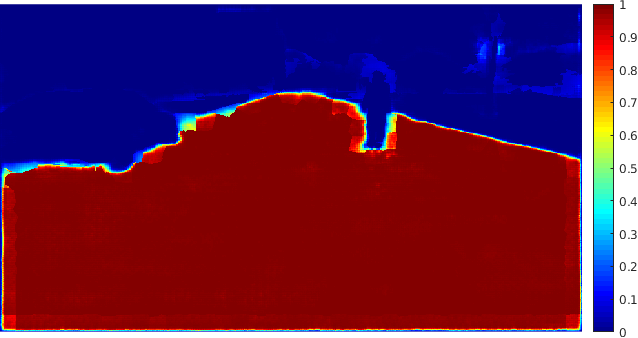}
    \caption{Road}
    \label{sub_q1}
    \end{subfigure}
\begin{subfigure}[]{0.32\columnwidth}
\centering
	\includegraphics[trim={0 0 2cm 0},clip,width=\columnwidth]{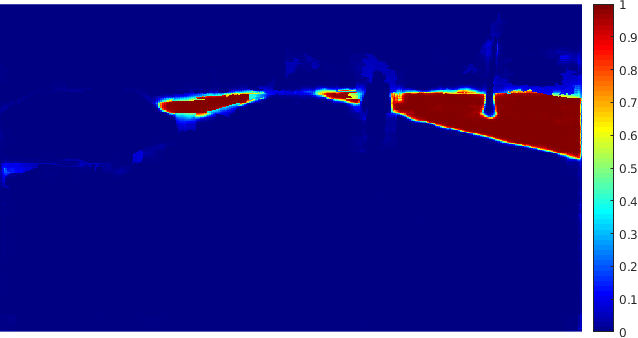}
    \caption{U. Road}
    \label{sub_r1}
    \end{subfigure}

\begin{subfigure}[]{0.32\columnwidth}
\centering
	\includegraphics[trim={0 0 2cm 0},clip,width=\columnwidth]{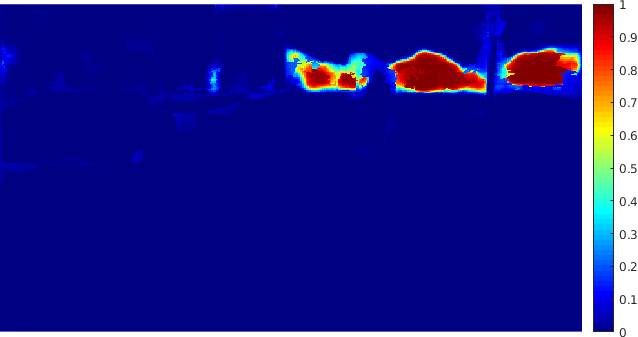}
    \caption{Vegetation}
    \label{sub_s1}
    \end{subfigure}
\begin{subfigure}[]{0.32\columnwidth}
\centering
	\includegraphics[trim={0 0 2cm 0},clip,width=\columnwidth]{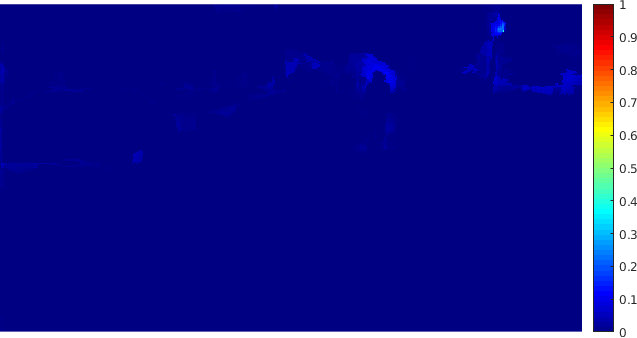}
    \caption{Sign}
    \label{sub_t1}
    \end{subfigure}
\begin{subfigure}[]{0.32\columnwidth}
\centering
	\includegraphics[trim={0 0 2cm 0},clip,width=\columnwidth]{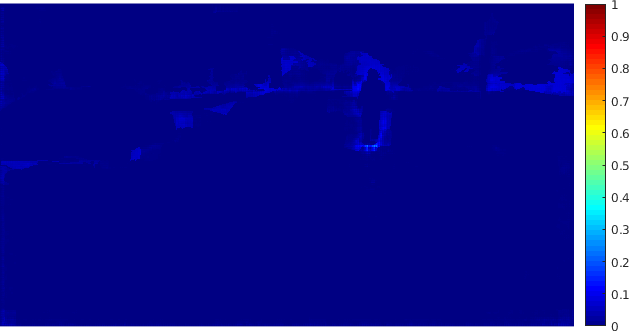}
    \caption{Fence}
    \label{sub_u1}
    \end{subfigure}
\begin{subfigure}[]{0.32\columnwidth}
\centering
	\includegraphics[trim={0 0 2cm 0},clip,width=\columnwidth]{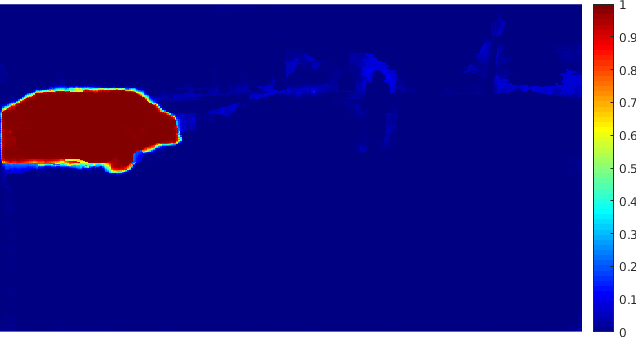}
    \caption{Vehicle}
    \label{sub_v1}
    \end{subfigure}
\begin{subfigure}[]{0.32\columnwidth}
\centering
	\includegraphics[trim={0 0 2cm 0},clip,width=\columnwidth]{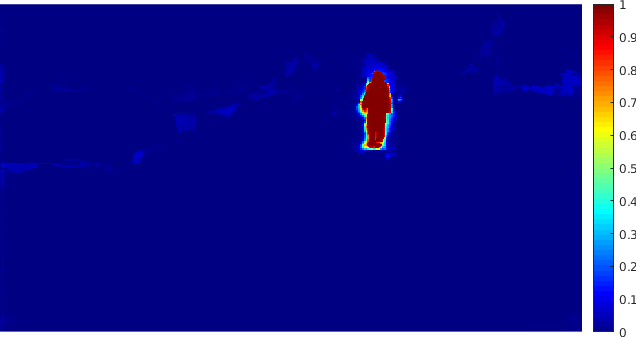}
    \caption{Pedestrian}
    \label{sub_w1}
    \end{subfigure}
\begin{subfigure}[]{0.32\columnwidth}
\centering
	\includegraphics[trim={0 0 2cm 0},clip, width=\columnwidth]{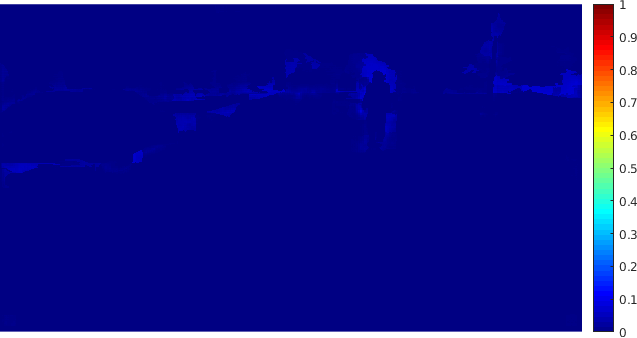}
    \caption{Rider}
    \label{sub_y1}
    \end{subfigure}

\caption{\small Uncertainty association process, continued. Comparison between the original CNN's score maps (first two rows) and the proposed method (last two rows) changing the temperature of the softmax function based on $spp_k$. Dark red represents the highest class probability and dark blue is the lowest class probability.}
\label{fig:proceso1}
\end{figure*}

The projection of the image information to the point cloud is done by applying a masking technique that accounts for the occlusion problem. This is achieved by discarding the lidar points which are not visible from the camera due to temporal and spatial offsets between the sensors. Each valid 3D point incorporates the probabilistic distribution of the semantic classes calculated in an earlier step.
The probabilistic semantic point cloud resulting from the previous processes is registered through a modified octomap framework. The different processes are presented in detail in the following sections.

\subsection{Heuristic label probability}

ENet's final module consists of a bare full convolution. This network outputs a three dimensional $c$ x $n$ x $m$ feature/activation map, where $n$ and $m$ correspond to the size of the input image, and $c$ is the number of object classes \cite{Paszke2016ENetAD}. 
Each $c$ feature map consists of activations per pixel $(u,v)$ which represents the unnormalised class score $S_c$ for each label \cite{simonyan14deep}. 

The output unit activation function for the CNN model represents the canonical link as determined by the softmax function (the most frequently used softmax method, assumes a Gibbs or Boltzmann distribution) \cite{10.1007/978-3-642-76153-9_28_softmax}:  

\begin{equation} \label{eq1}
P_c=\frac{\exp({S(c))}}{\sum_{b=1}^{n_c}\exp({S(b)})}.
\end{equation}

The softmax function mixes the class scores $S_c$ while satisfying the constraints, adopting the interpretation of $P_c$ as class probabilities \cite{Bishop:2006:PRM:1162264}:

\begin{equation} \label{eq2}
\sum_{c}P_c = 1, \qquad 0\leq P_c\leq 1.
\end{equation}

The final labeled image $L$ is a $n$ x $m$ matrix composed of the class identifier of the label with a higher probability per pixel.

In this paper, we propose a variant of the method for obtaining the label probabilities (while keeping the CNN's output classification) based on the score maps and the distribution of the label within segmented elements that compose the input image.
Initially, the input image is divided into different regions by using simple linear iterative clustering (SLIC) \cite{6205760_slic} super-pixels segmentation method \cite{1238308_superpixel}, where pixels are grouped concerning perceptual characteristics consistency. 

As presented in \cite{IROS2018_stephany}, we consider that due to the uniformity of the pixels within a super-pixel, these pixels are likely to belong to a unique class.
In order to measure the coherence with this last statement, we calculate the percentage of the predominant label $spp_k$ within the super-pixel $k$ dividing the number of pixels belonging to this label by the total amount of pixels inside the super-pixel. In an ideal case where the labels and the super-pixels are correctly segmented, the most prevalent label percentage $spp_k$ would be $1$. Fig. \ref{fig:proceso} shows the result of this process, it is noticeable that the value of $spp_k$ decreases in the super-pixels near the object borders.

Sometimes, two or more labels are contained within a super-pixel, indicating a number of instances where pixels were incorrectly labelled. This problem is in general due to the re-sizing process performed by the CNN model, which mainly affects the classification of the edges of objects within an image. In this case, we would expect to have a reasonably uniform probability distribution consisting of two or more labels. Nevertheless, the softmax function described in (\ref{eq1}) generates a strong discrepancy in the selection probability for dissimilar estimated class scores.

We have unified the concepts of predominant label percentage within a super-pixel, and the softmax activation function used by the CNN to select the most likely label per pixel. This is done using an alternative definition of the softmax function: 
\begin{equation} \label{eq3}
P_c=\frac{\exp({S(c)/\tau) }}{\sum_{b=1}^{n}\exp({S(b)/\tau })},
\end{equation}
where $\tau$ represents a positive parameter denominated as temperature.  In this case, high temperatures lead to the selection probability being approximately equiprobable. Low temperatures result in a greater difference in selection probability \cite{sutton1992reinforcement}. 

Our approach consists of modulating the temperature of the softmax function per super-pixel based on its label distribution.  With this approach, we obtain more coherent estimated probability of the classification process while satisfying the restriction (\ref{eq2}) and maintaining the classification output. The softmax temperature adjustment for the super-pixel $k$ is implemented as follows: 

\begin{equation} \label{eq4}
\tau_k=\frac{1}{spp_k^2}.
\end{equation}

The softmax temperature per super-pixel is inversely proportional to the square of $spp_k$, so, when $spp_k$ is less than $1$, the temperature is raised, flattening the activation function and consequently generating more distributed probabilities. Whereas $spp_k$ equals to 1, the class probabilities are identical to ones provided by the CNN model, as shown in Fig. \ref{fig:proceso1}.

\subsection{Calibration}

The cameras located on the vehicle have a lens of $100^\circ$ horizontal field of view, classified as a fisheye lens. We have calibrated the cameras using a variant of the ROS package \textit{camera calibration} \cite{ros_camera_calib} that uses a generic camera model \cite{fish_eye_model}. The camera intrinsic parameters for this model consist of the focal length, principal points and 4 fish-eye equidistant distortion coefficients. 
The extrinsic calibration was conducted as we specified in \cite{SurabhiITSC}. This process uses a checkerboard from which the same features are extracted by both the camera and the lidar. The features are the centre point and the normal vector of the board. These features are fed to a genetic algorithm which is in charge of optimizing the geometrical extrinsic parameters of the 3D transformation $T_{l}^{cn}$ between the two sensors.

\subsection{Motion correction and pixel coordinate calculation}  

The motion of the egocentric robot platform distorts the lidar measurements as the sensor coordinate system moves along with the platform during the period of a scan. In theory, every 3D point is measured from a temporally unique frame of reference. The lidar points, therefore, must be compensated for the motion of the platform before further point cloud and sensor fusion related processing is implemented.

\begin{figure}[!h]
\vspace{3mm}
\centerline{
\includegraphics[width=0.95\columnwidth]{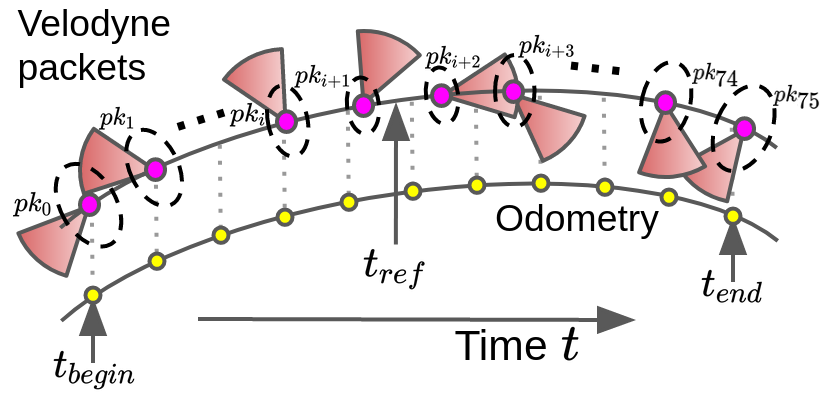}
}
\caption{\small lidar point cloud motion correction process. The Velodyne VLP-16 lidar used in our vehicle platform produces 76 lidar packets for each full revolution scan. The motion correction in this case is carried out for each packet with respect to the reference timestamp $t_{ref}$.}
\label{fig:time_line}
\end{figure}

The lidar measurements with similar timestamps are usually grouped into a single lidar packet, with a common timestamp assigned to the set of measurements for the convenience of processing. Each of the lidar packets is transformed based on the estimated delta translation and rotation of the vehicle platform between the packet timestamp and the image reference timestamp $t_{ref}$, as illustrated in Figure \ref{fig:time_line}. The proposed approach makes use of the unscented transform (UT) to propagate the uncertainties from the ego-motion estimation to the corrected 3D lidar points and then to projected pixel coordinates in each camera image. The entire process can be divided into three cascaded stages, namely vehicle ego-motion prediction, lidar motion correction, and lidar-to-camera projection, each can be included in the UT pipeline.

\begin{figure}[h!]
\vspace{3mm}
\centering

\begin{subfigure}[]{0.48\columnwidth}
\centering
	\includegraphics[width=\columnwidth]{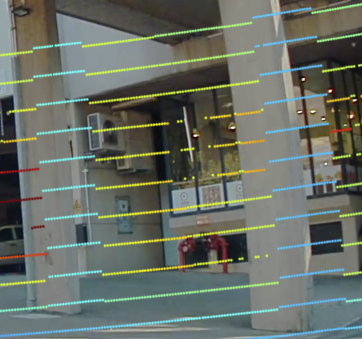}
    \caption{Before motion correction}
    \label{submc_a}
    \end{subfigure}
\begin{subfigure}[]{0.48\columnwidth}
\centering
	\includegraphics[width=\columnwidth]{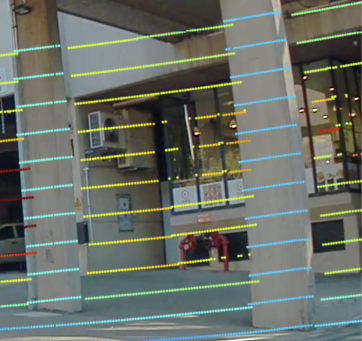}
    \caption{After motion correction}
    \label{submc_b}
    \end{subfigure}

\caption{\small Point cloud to image projection for raw and motion corrected lidar point cloud. }
\label{fig:MC_before_after}
\end{figure}

We assume a lidar scan is comprised of a set of $N$ lidar packets and their timestamps denoted as
\begin{equation}
\left\{pk_{i}, t_{i}^{pk}\right\}_{i=0}^{N-1},
\end{equation}
where $pk_{i}$ contains a set of $M$ 3D lidar measurement points $\left\{\bm{z}_{i,j}^{ld}\right\}_{j=0}^{M-1}$, and $\bm{z}^{ld} = \begin{bmatrix} x^{ld} & y^{ld} & z^{ld} & 1 \end{bmatrix}^{T}$.

The reference time $t_{ref}$ is usually chosen to be the timestamp corresponding to a common frame of reference where sensor fusion or subsequent processing is implemented. In scenarios where camera-lidar sensor fusion is desired, the rectification of the lidar points will have to be matched with the timestamp of the associated camera frame before the lidar-to-camera projection can be carried out \cite{8317846_supersensor}. For instance, the $t_{ref}$ can be set to coincide with the timestamp of the most recent or closest image.

Before we proceed, the UT state decomposition and recovery functions are presented in Table \ref{table:UT_Decompose} and Table \ref{table:UT_Restore} respectively for the convenience of subsequent discussion.  In these tables, $\lambda = \alpha^2\left(d+\kappa\right)-d$, $d = dim\left(\textbf{x}\right)$ is the dimension of state $\textbf{x}$, scaling parameters $\kappa \ge 0$, $\alpha \in \left(0, 1\right]$, and $\beta = 2$ for Gaussian distribution, $\left(\sqrt{\bm{\Sigma}}\right)_i$ is to obtain the $i^{th}$ column of the matrix square root $\textbf{R} = \sqrt{\bm{\Sigma}}$, which can be computed by Cholesky decomposition such that we have $\bm{\Sigma} = \textbf{R}\textbf{R}^{T}$.

\begin{table}[!t]
	\centering
	\caption{\small Algorithm: State Decomposition in Unscented Transform}
	\label{table:UT_Decompose}
	\scalebox{1.2}{
		\begin{tabular}{cl}
			\multicolumn{2}{l}{$\left\{\mathcal{X}_{i}, w_{i}^{m}, w_{i}^{c}\right\}_{i=0}^{2d} \leftarrow UTD\left(\bar{\textbf{x}}, \Sigma\right)$}\\
			1: & \hspace{-10pt}
			$\mathcal{X}_{0} = \bar{\textbf{x}}$
			\\
			2: & \hspace{-10pt}
			$\mathcal{X}_{i} = \bar{\textbf{x}} + \left(\sqrt{\left(d+\lambda\right)\Sigma}\right)_i\ \text{for}\ i=1,\cdots,d$
			\\
			3: & \hspace{-10pt}
			$\mathcal{X}_{i} = \bar{\textbf{x}} - \left(\sqrt{\left(d+\lambda\right)\Sigma}\right)_i\ \text{for}\ i=d+1,\cdots,2d$
			\\
			4: & \hspace{-10pt}
			$w_{0}^{m} = \frac{\lambda}{d+\lambda}$
			\\
			5: & \hspace{-10pt}
			$w_{0}^{c} = \frac{\lambda}{d+\lambda} + \left(1-\alpha^2+\beta\right)$
			\\
			6: & \hspace{-10pt}
			$w_{i}^{m} = w_{i}^{c} = \frac{1}{2\left(d+\lambda\right)}\ \text{for}\ i=1,\cdots,2d$
	\end{tabular}
	}
\end{table}

\begin{table}[!b]
	\centering
	\caption{\small Algorithm: State Recovery in Unscented Transform}
	\label{table:UT_Restore}
	\scalebox{1.2}{
		\begin{tabular}{cl}
			\multicolumn{2}{l}{$\bar{\textbf{x}}, \Sigma \leftarrow UTR\left(\left\{\mathcal{X}_{i}, w_{i}^{m}, w_{i}^{c}\right\}_{i=0}^{2d}\right)$}\\
			1: & \hspace{-10pt}
			$\bar{\textbf{x}} = \sum_{i=0}^{2d}{w_{i}^{m} \mathcal{X}_{i}}$
			\\
			2: & \hspace{-10pt}
			$\Sigma = \sum_{i=0}^{2d}{w_{i}^{c} \left(\mathcal{X}_{i}-\bar{\textbf{x}}\right) \left(\mathcal{X}_{i}-\bar{\textbf{x}}\right)^T}$
			\\
	\end{tabular}}
\end{table}

\subsubsection{Vehicle Ego-Motion Prediction}

To estimate the ego-motion of the moving vehicle it is necessary to predict the change in pose using observations from the environment. In the vehicle platform used for our experiments, the instantaneous linear and angular velocities are measured using on board wheel encoders and an IMU at a rate of 100 Hz. Based on a sequence of monotonically increasing packet timestamps $\bm{t}_{0:N-1}^{pk} = \left\{t_{i}^{pk}\right\}_{i=0}^{N-1}$, it is reasonable to construct a sequence of linear velocity vectors $\bm{z}_{0:N-1}^{v} = \left\{\bm{z}_{i}^{v}\right\}_{i=0}^{N-1}$ corresponding to $\bm{t}_{0:N-1}^{pk}$, and likewise a sequence of angular velocity vectors $\bm{z}_{0:N-1}^{\omega} = \left\{\bm{z}_{i}^{\omega}\right\}_{i=0}^{N-1}$.

Each $\bm{z}_{i}^{v}$ is a column vector with linear velocity readings along with $x$, $y$, and $z$ and each $\bm{z}_{i}^{\omega}$ is a column vector with the angular velocity measurements in $roll$, $pitch$, and $yaw$ in the local frame of reference of the vehicle. In cases where odometry data and lidar packets are asynchronous, $\bm{z}_{i}^{v}$ and $\bm{z}_{i}^{\omega}$ can be approximated using those with the closest timestamps to $t_{i}^{pk}$, respectively, assuming that the vehicle velocity change during the time difference is negligible. Also, $\bm{z}_{i}^{v}$ and $\bm{z}_{i}^{\omega}$ are assumed to contain independently and identically distributed zero-mean Gaussian noises with their covariance matrices denoted as $\bm{\Sigma}^{v}$ and $\bm{\Sigma}^{\omega}$, respectively. The timing jitter in $t_{i}^{pk}$ is modelled as zero-mean Gaussian noise with standard deviation $\sigma_{t}$.

Given $\bm{t}_{0:N-1}^{pk}$, $\bm{z}_{0:N-1}^{v}$, $\bm{z}_{0:N-1}^{\omega}$, and $t_{ref}$, the vehicle ego-motion prediction is estimate a sequence of Gaussian variables $\left\{\textbf{x}_{i}^{veh} \sim \mathcal{N}\left(\bar{\textbf{x}}_{i}^{veh}, \bm{\Sigma}_{i}^{veh}\right)\right\}_{i=0}^{N-1}$ representing the predicted egocentric vehicle poses at $t^{pk}$ with respect to that at $t_{ref}$. Let $\textbf{x}_{ref}^{veh}$ denote the Gaussian variable representing the vehicle egocentric state at $t_{ref}$.
\begin{equation}
\textbf{x}_{ref}^{veh} \sim \mathcal{N}\left(\bar{\textbf{x}}_{ref}^{veh}, \Sigma_{ref}^{veh}\right),
\end{equation}
where we initialize the mean vector $\bar{\textbf{x}}_{ref}^{veh} = 0$ and the covariance matrix $\bm{\Sigma}_{ref}^{veh}$ to a diagonal matrix with close to zero elements, since we are performing ego-motion prediction within the local coordinate system of the vehicle at $t_{ref}$.

If $t_{ref} > t_{0}^{pk}$, then reverse vehicle ego-motion prediction is performed by first initialising intermediate variables:
\begin{align} \label{eq:egomotion_prediction_init}
t_{*} &\leftarrow t_{ref} & \bar{\textbf{x}}_{*}^{veh} &\leftarrow \bar{\textbf{x}}_{ref}^{veh} & \bm{\Sigma}_{*}^{veh} &\leftarrow \bm{\Sigma}_{ref}^{veh}.
\end{align}

Then for $i = \max\left( p : t_{p}^{pk} \in \bm{t}^{pk} \wedge t_{p}^{pk} < t_{ref} \right), \cdots, 1, 0$, an augmented state vector is constructed by concatenating the intermediate vehicle egocentric state $\textbf{x}_{*}^{veh}$ and velocity measurements at $t_{i}^{pk}$.

\begin{equation} \label{eq:egomotion_prediction_start}
\textbf{x}_{*}^{a}
\sim
\mathcal{N}\left(
\bar{\textbf{x}}_{*}^{a},
\bm{\Sigma}_{*}^{a}
\right),
\end{equation}
where
\(
\bar{\textbf{x}}_{*}^{a} =
\begin{bmatrix}
\left(\bar{\textbf{x}}_{*}^{veh}\right)^{T} & \left(\bm{z}_{i}^{v}\right)^{T} & \left(\bm{z}_{i}^{\omega}\right)^{T} & t_{i}^{pk} & t_{*}
\end{bmatrix}
\), and
$\bm{\Sigma}_{*}^{a} = \text{blkdiag}\left\{\bm{\Sigma}_{*}^{veh}, \bm{\Sigma}_{v}, \bm{\Sigma}_{\omega}, \sigma_{t}^{2}, \sigma_{t}^{2}\right\}$.

The backward motion prediction goes from a later timestamp $t_{*}$ to an earlier $t_{i}^{pk}$, resulting in a negative value in time difference considered in the kinematic model.

The augmented state mean and covariance matrix are UT decomposed into a set of sigma points.
\begin{equation}
\left\{\mathcal{X}_{j}^{a}, w_{j}^{m}, w_{j}^{c}\right\}_{j=0}^{2d} \leftarrow UTD\left(\bar{\textbf{x}}_{*}^{a}, \bm{\Sigma}_{*}^{a}\right)
\end{equation}

For $j = 0,\cdots,2d$, motion prediction is conducted backward in time.
\begin{equation}
\mathcal{Y}_{j}^{veh} = f_{km}\left(\mathcal{X}_{j}^{a}\right),
\end{equation}
where $f_{km}\left(\cdot\right)$ is the vehicle kinematic model that predicts vehicle pose based on a given pose and kinematic velocities over a given time duration.

The predicted vehicle egocentric state at timestamp $t_{i}^{pk}$ is recovered by
\begin{equation}
\bar{\textbf{x}}_{i}^{veh}, \bm{\Sigma}_{i}^{veh} \leftarrow UTR\left(\left\{\mathcal{Y}_{j}^{veh}, w_{j}^{m}, w_{j}^{c}\right\}_{j=0}^{2d}\right).
\end{equation}

The results also serve as intermediate variables for the next iteration:
\begin{align} \label{eq:egomotion_prediction_end}
t_{*} &\leftarrow t_{i}^{pk} & \bar{\textbf{x}}_{*}^{veh} &\leftarrow \bar{\textbf{x}}_{i}^{veh} & \bm{\Sigma}_{*}^{veh} &\leftarrow \bm{\Sigma}_{i}^{veh}.
\end{align}

If \(t_{ref} \leq t_{N-1}^{pk}\), then forward vehicle ego-motion prediction is carried out by initialising intermediate variables as in \eqref{eq:egomotion_prediction_init}, and for \(i = \min\left(p : t_{p}^{pk} \in \bm{t}^{pk} \wedge t_{p}^{pk} \geq t_{ref}\right),\cdots,N-1\), using the same set of equations \eqref{eq:egomotion_prediction_start}-\eqref{eq:egomotion_prediction_end}, except that in this case
\(
\bar{\textbf{x}}_{*}^{a} =
\begin{bmatrix}
\left(\bar{\textbf{x}}_{*}^{veh}\right)^{T} & \left(\bm{z}_{i}^{v}\right)^{T} & \left(\bm{z}_{i}^{\omega}\right)^{T} & t_{*} & t_{i}^{pk}
\end{bmatrix}
\), and in every iteration the motion estimation starts from an earlier timestamp $t_{i}^{pk}$ to a later $t_{*}$.

\subsubsection{3D Lidar Points Motion Correction}

With a sequence of predicted egocentric vehicle poses $\left\{\textbf{x}_{i}^{veh} \sim \mathcal{N}\left(\bar{\textbf{x}}_{i}^{veh}, \bm{\Sigma}_{i}^{veh}\right)\right\}_{i=0}^{N-1}$ at $\bm{t}^{pk}$ obtained from the vehicle ego-motion prediction stage, motion correction is applied to each corresponding packet of 3D lidar measurement points.

For $i = 0,1,\cdots,N-1$, the predicted state mean and covariance matrix are UT decomposed into a set of sigma points.
\begin{equation}
\left\{\mathcal{X}_{i,k}^{veh}, w_{i,k}^{m}, w_{i,k}^{c}\right\}_{k=0}^{2d} \leftarrow UTD\left(\bar{\textbf{x}}_{i}^{veh}, \bm{\Sigma}_{i}^{veh}\right)
\end{equation}

A set of $4\times 4$ homogeneous transformation matrices $\left\{\mathcal{T}_{i,k}^{veh}\right\}_{k=0}^{2d}$ are constructed based on the rotation and translation in each $\mathcal{X}_{i,k}^{veh}$.

For $j = 0,\cdots,M-1$, and for $k = 0,\cdots,2d$, a motion corrected sigma point is calculated as
\begin{equation}
\bm{\mathcal{Z}}_{i,j,k}^{cld} = \left(T_{veh}^{ld}\right)^{-1} \cdot \mathcal{T}_{i,k}^{veh} \cdot T_{veh}^{ld} \cdot z_{i,j}^{ld}
\end{equation}
where the lidar point is translated to the vehicle's base frame by the rigid transform $T_{veh}^{ld}$, followed by transformation that encapsulates delta ego-motion in vehicle base frame. Lastly, the point is translated back to the lidar coordinate system.

At this stage, a motion corrected lidar point $\bm{z}_{i,j}^{cld}$ within lidar packet $pk_{i}$ can be recovered to a Gaussian variable through
\begin{equation}
\bar{\bm{z}}_{i,j}^{cld}, \bm{\Sigma}_{i,j}^{cld} \leftarrow UTR\left(\left\{\bm{\mathcal{Z}}_{i,j,k}^{cld}, w_{i,k}^{m}, w_{i,k}^{c}\right\}_{k=0}^{2d}\right).
\end{equation}

Finally, the process produces a set of motion corrected sigma points for lidar points denoted by 
\begin{equation}
\bm{\Omega} = \left\{\left\{\left\{\bm{\mathcal{Z}}_{i,j,k}^{cld}\right\}_{j=0}^{M-1}, w_{i,k}^{m}, w_{i,k}^{c}\right\}_{k=0}^{2d}\right\}_{i=0}^{N-1}.
\end{equation}

This section described a process for lidar ego-motion correction with uncertainty. Further transformation can be applied to $\bm{\Omega}$ for constructing a lidar-to-camera projection with motion uncertainty. This is presented in the next section.


\subsubsection{Lidar-to-Camera Projection}


This component takes a motion-corrected 3D lidar point cloud as input and projects the points to a given camera image coordinate system. In the previous motion correction process, the timestamp of the image was used as the reference $t_{ref}$.
This was essential as now the pointcloud is synchronised to the image given vehicle motion.

Before the projection is applied, each 3D lidar point needs to be translated from the lidar frame to the camera frame given the extrinsic calibration between the camera and lidar sensor represented as the transformation matrix $T_{cam}^{ld}$:

\begin{equation}
\bm{z}^{cam} = T_{cam}^{ld} \bm{z}^{ld},
\end{equation}
where $\bm{z}^{cam} = \begin{bmatrix} x^{cam} & y^{cam} & z^{cam} & 1 \end{bmatrix}^{T}$ is the 3D lidar point translated to camera frame.

The generic lidar-camera projection function (see \ref{lidar-camera-fnc}ppendix) is defined as

\begin{equation}
\begin{bmatrix} u \\ v \end{bmatrix} =
f_{proj}
\left(\bm{z}^{cam}\right),
\end{equation}
which converts the pixel coordinates $u$ and $v$ into the image frame corresponding to a 3D lidar point $\bm{z}^{cam}$ in the camera frame by using the camera model and its intrinsic parameters.











Once the relationship between the 2D image coordinates and 3D lidar point is known, the information can be transferred from lidar to image domain and vice versa. For instance, the value of the semantic label is retrieved from the corresponding pixel and included into the point cloud as a descriptor for each 3D point.

In order to obtain projected points in the image frame with uncertainty information, and also to avoid unnecessary UT, this stage operates directly on $\bm{\Omega}$ which is the set of sigma points for each 3D lidar point as the output of the lidar motion correction stage.

We can combine the translation of each sigma point $\bm{\mathcal{Z}}_{i,j,k}^{cld} \in \bm{\Omega}$ from lidar frame to camera frame using $T_{cam}^{ld}$ and the projection to the image frame by

\begin{equation} \label{eq_pc0}
\left\{\bm{\mathcal{K}}_{i,j,k}^{cam} : \left(\exists \bm{\mathcal{Z}}_{i,j,k}^{cld} \in \Omega\right)
\left[
\bm{\mathcal{K}}_{i,j,k}^{cam} = f_{proj}\left(T_{cam}^{ld} \bm{\mathcal{Z}}_{i,j,k}^{cld}\right)
\right]
\right\}
\end{equation}

For $i = 0,\cdots,N-1$ and for $j = 0,\cdots,M-1$, the image pixel projected from the lidar point $\bm{z}_{i,j}^{cld}$ within lidar packet $pk_{i}$ can be recovered from the mean values and covariance matrix by
\begin{equation}
\begin{bmatrix} \bar{u}_{i,j} \\ \bar{v}_{i,j} \end{bmatrix}, \bm{\Sigma}_{i,j}^{uv} \leftarrow UTR\left(\left\{\bm{\mathcal{K}}_{i,j,k}^{cam}, w_{i,k}^{m}, w_{i,k}^{c}\right\}_{k=0}^{2d}\right)
\end{equation}

\subsection{Image-lidar projection}

The direct fusion of the camera-lidar information would be appropriate if both the camera and the lidar coordinate system were in the same location. Most vehicles are equipped with multiple cameras at different positions to achieve a wide coverage of the area of operation.  In this case, the cameras and lidar will see the environment from different vantage points. This fact introduces an issue where the lidar can observe objects that are behind objects that obstruct the visibility of the camera. This issue must be resolved before it is possible to fuse the camera and lidar information. A contribution of this paper is a new process to address the occlusion problem by applying a masking technique.

\subsubsection{Occlusion-handling}

In this section, we propose an efficient method to overcome the occlusion problem when projecting a sparse point cloud into an image where the sensors are not colocated.

Before the projection to the image frame, the point cloud in camera frame $[x^{cam},  y^{cam},  z^{cam}]$ is sorted using a k-dimensional tree. The 3D points are organized in ascending order, based on their distance to the camera frame origin, with the first point being the closest and the last one the furthest. To cope with the occlusion problem, we propose a masking approach, where every projected point generates a mask which will prevent other points from being placed within the masked zone on the image. 

\begin{figure}[t!]
\vspace{3mm}
\centering

\begin{subfigure}[]{0.49\columnwidth}
\centering
	\includegraphics[width=\columnwidth]{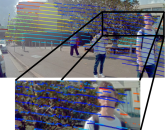}
    \caption{LS camera image.}
    \label{sub1_a}
    \end{subfigure}
\begin{subfigure}[]{0.49\columnwidth}
\centering
	\includegraphics[width=\columnwidth]{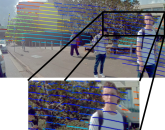}
    \caption{LS masked camera image. }
    \label{sub1_b}
    \end{subfigure}
    
\begin{subfigure}[]{0.49\columnwidth}
\centering
	\includegraphics[width=\columnwidth]{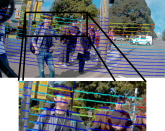}
    \caption{L camera image.}
    \label{sub1_c}
    \end{subfigure}
\begin{subfigure}[]{0.49\columnwidth}
\centering
	\includegraphics[width=\columnwidth]{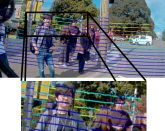}
    \caption{L masked camera image. }
    \label{sub1_d}
    \end{subfigure}
    
\begin{subfigure}[]{0.49\columnwidth}
\centering
	\includegraphics[width=\columnwidth]{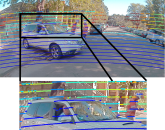}
    \caption{C camera image.}
    \label{sub1_e}
    \end{subfigure}
\begin{subfigure}[]{0.49\columnwidth}
\centering
	\includegraphics[width=\columnwidth]{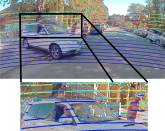}
    \caption{C masked camera image. }
    \label{sub2_a}
    \end{subfigure}
    
\begin{subfigure}[]{0.49\columnwidth}
\centering
	\includegraphics[width=\columnwidth]{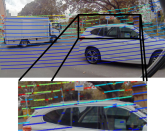}
    \caption{R camera image.}
    \label{sub2_b}
    \end{subfigure}
\begin{subfigure}[]{0.49\columnwidth}
\centering
	\includegraphics[width=\columnwidth]{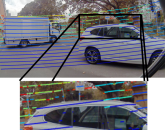}
    \caption{R masked camera image.}
    \label{sub2_c}
    \end{subfigure}
    
\begin{subfigure}[]{0.49\columnwidth}
\centering
	\includegraphics[width=\columnwidth]{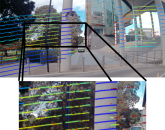}
    \caption{SR camera image. }
    \label{sub2_d}
    \end{subfigure}
\begin{subfigure}[]{0.49\columnwidth}
\centering
	\includegraphics[width=\columnwidth]{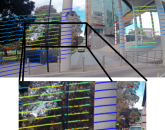}
    \caption{SR masked camera image.}
    \label{sub2_e}
    \end{subfigure}

\caption{\small Comparison between direct point cloud projection (first column) and after applying the occlusion masking technique (second column) for each camera on the vehicle: left side (LS),  left (L), centre (C), right (R), right side (RS)}
\label{fig:masked_images}
\end{figure}

The mask shape corresponds to a rectangle where the dimensions $x_{Gap}$ and $y_{Gap}$ depend on the vertical and horizontal resolution of the lidar. Based on the VLP-16 specifications \cite{velodyne}, the gap between points (vertical or horizontal) can be calculated with the following equation:
\begin{equation} \label{eq12}
    Gap=d_t*\tan(\theta_{v-h})
\end{equation}
where $d_t$ is the distance to the target and $\theta_{v-h}$ the vertical or horizontal angle between scan lines or consecutive points. In order to obtain the gaps in pixels when the point cloud is projected, we make use of the generic camera-image projection equations \ref{eq_pc1} and \ref{eq_pc4}, obtaining:

\begin{align} \label{eq13}
u_{Gap} &= f_x \frac{d_t\tan(\theta_h)}{z} &   v_{Gap} &= f_y \frac{d_t\tan(\theta_v)}{z}
\end{align}

\begin{table*}[t]
\centering
\caption{\small Single scan evaluation}
\label{table:SingleEvaluation}
\begin{tabular}{|c|c|c|c|c|c|c|c|c|c|}
\hline
\multirow{2}{*}{\textbf{Semantic class}} & \multicolumn{3}{c|}{\textbf{Direct Projection}} & \multicolumn{3}{c|}{\textbf{Projection + Motion Correction}} & \multicolumn{3}{c|}{\textbf{Projection + Motion C + Mask}} \\ \cline{2-10} 
                                & \textbf{Recall}  & \textbf{Precision } & \textbf{F1 Score} & \textbf{Recall}    & \textbf{Precision }    & \textbf{F1 Score}   & \textbf{Recall}  & \textbf{Precision }  & \textbf{F1 Score}  \\ \hline
Building                        & 0.752       & 0.786      & 0.769       & 0.771         & 0.805         & 0.788         & 0.811       & 0.852       & 0.830        \\ \hline
Pole                            & 0.706       & 0.174      & 0.279       & 0.725         & 0.187         & 0.297         & 0.732       & 0.218       & 0.336       \\ \hline
Road                            & 0.960       & 0.948      & 0.954       & 0.965         & 0.950         & 0.957         & 0.965       & 0.958       & 0.961        \\ \hline
Undrivable Road                         & 0.775       & 0.658      & 0.712       & 0.788         & 0.678         & 0.729         & 0.826       & 0.730       & 0.775        \\ \hline
Vegetation                      & 0.865       & 0.937      & 0.900       & 0.878         & 0.949         & 0.912         & 0.901       & 0.972       & 0.935        \\ \hline
Vehicle                         & 0.944       & 0.830      & 0.883       & 0.956         & 0.836         & 0.891         & 0.964       & 0.849       & 0.903        \\ \hline
Pedestrian                      & 0.941       & 0.365      & 0.526       & 0.955         & 0.377         & 0.540         & 0.989       & 0.652       & 0.785        \\ \hline
\end{tabular}
\end{table*}

For our case, we have assumed that the difference between $d_t$ and $z$ is negligible since the distances between the lidar and the cameras is significantly smaller than the distance from the vehicle and the obstacles, making $d_t \approx  z$. Therefore, $u_{Gap}$ and $v_{Gap}$ could be computed in terms of camera intrinsic parameters and lidar angular resolution:

\begin{align} \label{eq14}
u_{Gap} &= f_x \tan(\theta_h) & v_{Gap} &= f_y \tan(\theta_v)   
\end{align}

In the case of the VLP-16 lidar and our GMSL cameras, $\theta_v = 2 ^\circ$ and $\theta_h = 0.1 ^\circ$, therefore, $y_{Gap}= 41$ pixels and $x_{Gap} = 3$ pixels. Having calculated the size in pixels of the vertical and horizontal gaps, we proceed to project the ordered point cloud into the image, starting with the closest point to the camera. Each projected point will create a rectangular mask of $x_{Gap}$ and $y_{Gap}$ dimensions centred in the corresponding pixel. A 3D point will be considered occluded and won't be taken into account if its projection lies inside a masked area.
Fig. \ref{fig:masked_images} shows the difference between the lidar-image projection with and without the occlusion masking for each camera image.

\subsubsection{Label uncertainties projection}

With the valid (unoccluded) points, given $(u, v)$ coordinates in the corresponding camera image frame, their covariance matrix $\Sigma _{uv}$ and the heuristic probabilities computed for each label per single image, we can calculate the label distribution for each 3D point.

\begin{figure}[!h]
\vspace{3mm}
\centerline{
\includegraphics[width=0.8\columnwidth]{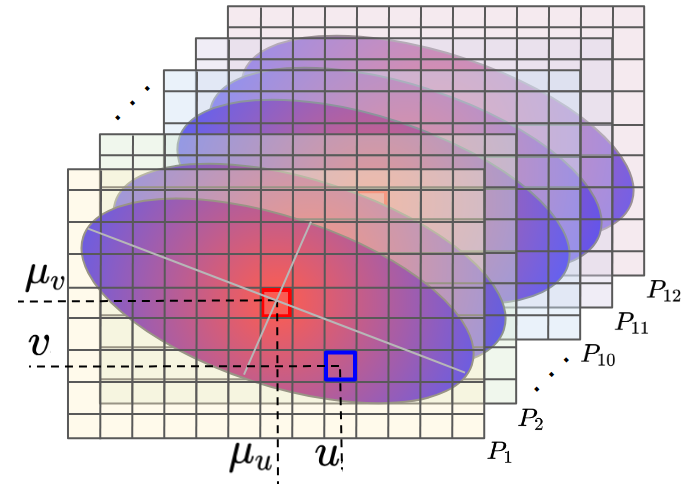}
}
\caption{\small Semantic class probability distribution projection.}
\label{fig:uncertainties}
\end{figure}

Given a 3D point $(x_i,y_i,z_i)$, we obtain its semantic label probability $L_{ic}$ for each of the classes $c$ as: 

\begin{align} \label{eq15}
L_{ic}= \eta \sum_{u=\mu_u-u_p}^{\mu_u+u_p} \sum_{v=\mu_v-v_p}^{\mu_v+v_p} P_c(u,v)*f(u,v,\Sigma _{uv})
\end{align}

where $u_p$ and $v_p$ are the limits of the ellipse, which represents 90\% confidence in the image frame. $\eta$ is a normalization factor which assures that the sum of all $L_{ic}$ equals to 1, and the function $f$ corresponds to the bivariate probability density function.

 As a result of this process, each 3D point will get a list of probabilities of belonging to a particular class. These probabilities are evaluated as a function of uncertainties in the labeling and motion correction processes. We consider these new point cloud attributes essential to get a better understanding of the current state of the world.

\subsection{Point cloud registration}

We used an Octomap framework (octree representation of the environment) to register the resulting point cloud. The Octomap representation was adapted to update the occupancy probability of each voxel. It is also appealing to represent and update the information of the different semantic classes. Each 3D unit is represented with its coordinates, occupancy and semantic probability distribution. 

Each time a new semantic 3D point is registered, it is initially translated from the vehicle to the map frame. Given the position of the lidar in the map frame, a ray casting operation is performed to clear all the voxels between the sensor origin and the lidar endpoint to initialize or update the voxel it occupies. The occupancy probability of each voxel is updated by a binary bayes filter, in which a discrete bayes filter algorithm updates the semantic class label probabilities. 

\section{Results}

We tested our algorithm on the Usyd Dataset \cite{dataset_paper}. An electric vehicle equipped with multiple sensors was driven around the University of Sydney. The following information was collected: images (from five GSML $100^\circ$ field of view cameras located around the platform), point clouds (from one lidar VLP-16 located on the roof), and odometry information from an inertial measurement unit IMU and wheel encoders.

\begin{figure*}[h!]
\vspace{3mm}
\centering

\begin{subfigure}[]{1.0\textwidth}
\centering
	\includegraphics[width=0.9\textwidth]{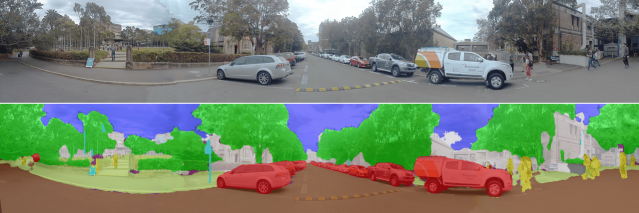}
    \caption{Stitched images for a 350$^\circ$ field of view of the surroundings. }
    \label{sub_es}
    \end{subfigure}

\begin{subfigure}[]{0.66\columnwidth}
\centering
	\includegraphics[width=0.86\columnwidth]{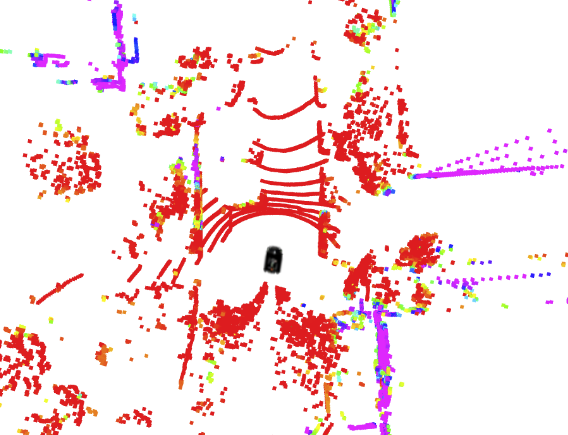}
    \caption{Buildings.}
    \label{sub_p_b}
    \end{subfigure}
\begin{subfigure}[]{0.66\columnwidth}
\centering
	\includegraphics[width=0.86\columnwidth]{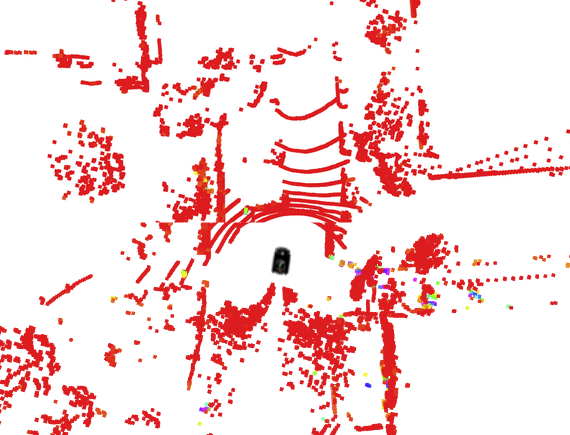}
    \caption{Poles and pedestrians. }
    \label{sub_p_pp}
    \end{subfigure}
\begin{subfigure}[]{0.66\columnwidth}
\centering
	\includegraphics[width=0.86\columnwidth]{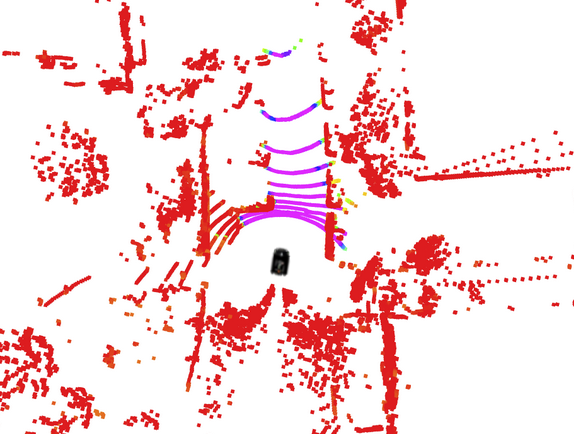}
    \caption{Road.}
    \label{sub_p_r}
    \end{subfigure}
    
\begin{subfigure}[]{0.66\columnwidth}
\centering
	\includegraphics[width=0.86\columnwidth]{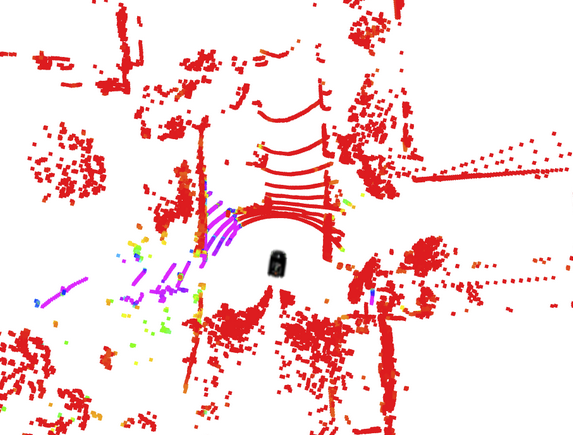}
    \caption{Undrivable road.}
    \label{sub_p_ur}
    \end{subfigure}
\begin{subfigure}[]{0.66\columnwidth}
\centering
	\includegraphics[width=0.86\columnwidth]{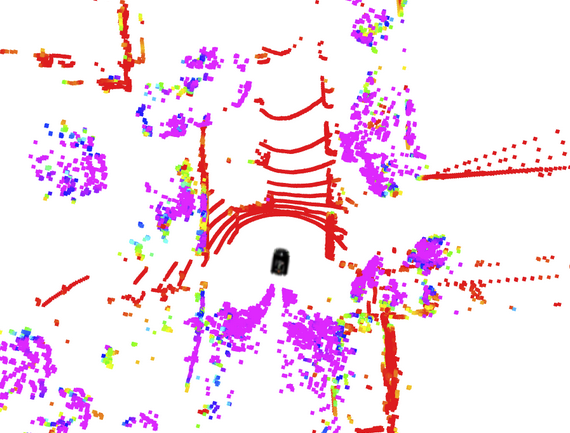}
    \caption{Vegetation. }
    \label{sub_p_v}
    \end{subfigure}
\begin{subfigure}[]{0.66\columnwidth}
\centering
	\includegraphics[width=0.86\columnwidth]{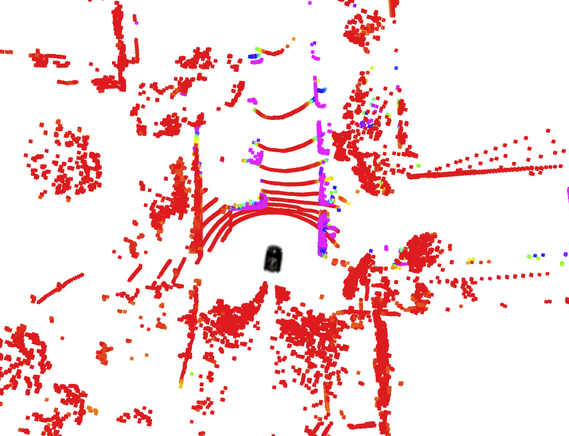}
    \caption{Vehicles.}
    \label{sub_p_vh}
    \end{subfigure}

\begin{subfigure}[]{\textwidth}
\centering
	\includegraphics[width=0.99\textwidth, trim={0 4cm 0  4cm},clip]{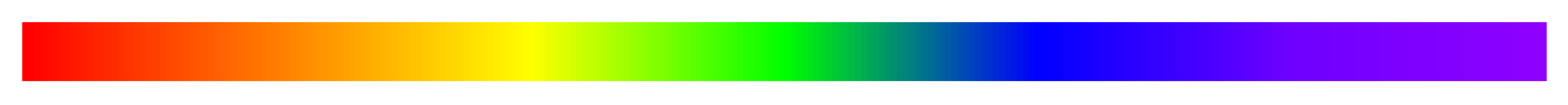}
    \end{subfigure}

\caption{\small \ref{sub_es} Semantic segmentation of images from the surroundings. Single-scan point cloud with probabilities for each semantic class with red color for the lowest probability $(0.0)$ and violet the highest $(1.0)$.
In \ref{sub_p_b} is possible to see in purple at the right-bottom a structure forming a 90$^\circ$ angle corresponding to a building, while in the next image scattered green, blue and purple points represent poles and pedestrians around the vehicle. Directly in front of the vehicle in \ref{sub_p_r}, the purple section indicates the part of the road visible to the camera, on its left in \ref{sub_p_ur}, there's a portion of undrivable road. In the following images \ref{sub_p_v} and \ref{sub_p_vh}, vegetation and vehicles (on the side of the road) are coloured in purple. 
}
\label{fig:six_uncert}
\end{figure*}

To assess the performance of the proposed pipeline for a single laser scan, we compare three different strategies for the lidar-image projection. Firstly, we directly project the semantic labels from the image to the closest point in the point cloud. The second strategy incorporates motion correction to compensate for the vehicle movement during the lidar scan. The final strategy incorporates both the occlusion handling method presented in this paper along with the motion correction. 

A total of 20 single lidar scans were hand labelled for this experiment, and are used as the ground truth to measure the performance of the various processing strategies. In the second and third strategies evaluated here, the information projected into the point cloud results in a semantic class probability distribution.  For these approaches, the performance is measured by comparing the ground truth and the semantic class with the highest probability. 

We combined and discarded unused semantic classes obtaining 7 final labels for the point cloud (\textit{sky} and \textit{unlabeled} classes were discarded while \textit{pole} and \textit{sign}, \textit{pedestrian} and \textit{rider} and, \textit{building} and \textit{fence} were merged). We calculated the recall, precision and F1 score, the results of which are shown in Table \ref{table:SingleEvaluation}.

Overall, recall and precision metrics improved with the inclusion of motion correction and occlusion handling into the projection process. 
The number of labelled points per scan is around 7$\%$ less after the occlusion handling process compared to the direct projection.  The recall is improved by introducing this technique as it avoids occluded points that would otherwise be mislabelled, reducing the number of both false negatives and false positives.
Similarly, the correction of the point cloud for motion leads to a more accurate transfer of the labels as there is less error in the projection when taking into account the true vehicle position, and from correcting the point cloud to the specific image timestamp. 

\begin{table*}[t]
\centering
\caption{\small Scan registration evaluation}
\label{table:MapEvaluation}
\begin{tabular}{|c|c|c|c|c|c|c|c|c|c|}
\hline
\multirow{2}{*}{\textbf{Semantic Class}} & \multicolumn{3}{c|}{\textbf{Direct Projection}} & \multicolumn{3}{c|}{\textbf{Projection + Motion Correction}} & \multicolumn{3}{c|}{\textbf{Projection + Motion C + Mask}} \\ \cline{2-10} 
                                & \textbf{Recall}  & \textbf{Precision} & \textbf{F1 Score} & \textbf{Recall}    & \textbf{Precision}    & \textbf{F1 Score}   & \textbf{Recall}  & \textbf{Precision}  & \textbf{F1 Score}  \\ \hline
Building                        & 0.729       & 0.875      & 0.795       & 0.768         & 0.907         & 0.831         & 0.769       & 0.923       & 0.839        \\ \hline
Pole                            & 0.632       & 0.195      & 0.298       & 0.648         & 0.213         & 0.320         & 0.681       & 0.248       & 0.364        \\ \hline
Road                            & 0.722       & 0.921      & 0.809       & 0.732         & 0.923         & 0.816         & 0.734       & 0.945       & 0.826        \\ \hline
U. Road                         & 0.347       & 0.529      & 0.419       & 0.375         & 0.540         & 0.443         & 0.415       & 0.580       & 0.485        \\ \hline
Vegetation                      & 0.910       & 0.848      & 0.878       & 0.918         & 0.854         & 0.885         & 0.920       & 0.879       & 0.899        \\ \hline
Vehicle                         & 0.948       & 0.525      & 0.676       & 0.960         & 0.532         & 0.685         & 0.970       & 0.558       & 0.708        \\ \hline
Pedestrian                      & 0.525       & 0.166      & 0.252       & 0.537         & 0.176         & 0.265         & 0.553       & 0.204       & 0.298        \\ \hline
\end{tabular}
\end{table*}

From Table \ref{table:SingleEvaluation}, it is evident that the semantic classes showing the greatest improvement in the performance metrics are poles and pedestrians. Poles are generally skinny, a lidar such as the VLP 16 has a low vertical resolution, so few points are detected from the object. If the projection is shifted due to vehicle motion, the transferred labels could be offset by more than the width of the object making the result completely incorrect.
Pedestrians are typically moving objects; the synchronisation between the point cloud and the image plays a fundamental role in the overlay of the sensor information.
Likewise, the position of the objects is relevant when transferring accurately the semantic labels. Objects that are closer to the platform occlude other more distant objects which cause a mislabelling of the class, increasing the number of false positives. As poles and pedestrians are generally in the foreground of an urban road environment, these classes generally have the lowest precision and improve the most from the proposed occlusion handling method. 

\begin{table}[b]
\centering
\caption{\small Normalized Confusion Matrix: semantic single scan built with our pipeline (projection + motion correction + mask)}
\label{table:CM_single_scan}
\begin{tabular}{|l|c|c|c|c|c|c|c|c|}
\hline
\multicolumn{2}{|l|}{\multirow{2}{*}{}}                 & \multicolumn{7}{c|}{\textbf{True Label}}                                                   \\ \cline{3-9} 
\multicolumn{2}{|l|}{}                                  & \textbf{B} & \textbf{P} & \textbf{R} & \textbf{U} & \textbf{V} & \textbf{Vh} & \textbf{Pe} \\ \hline
\multirow{7}{*}{\rotatebox[origin=c]{90}{\textbf{Predicted Label}}} & \textbf{B}  &\cellcolor[HTML]{cc33ff} 81.1       & \cellcolor[HTML]{f9e6ff}8.3        & 0.0        & 1.6        & \cellcolor[HTML]{f9e6ff}6.4        & 1.2         & 0.2         \\ \cline{2-9} 
                                          & \textbf{P}  & 1.0        & \cellcolor[HTML]{d24dff}73.2       & 0.1        & 1.3        & 0.7        & 0.1         & 0.0         \\ \cline{2-9} 
                                          & \textbf{R}  & 0.5        & 2.7        & \cellcolor[HTML]{bf00ff}96.5       & \cellcolor[HTML]{f9e6ff} 7.9        & 0.4        & 1.4         & 0.0         \\ \cline{2-9} 
                                          & \textbf{U}  & 3.3        & 2.7        & 0.4        & \cellcolor[HTML]{cc33ff}82.6       & 0.8        & 0.4         & 0.4         \\ \cline{2-9} 
                                          & \textbf{V}  &\cellcolor[HTML]{f9e6ff} 8.2        & \cellcolor[HTML]{f9e6ff}9.8        & 0.2        & 0.5        & \cellcolor[HTML]{bf00ff} 90.1       & 0.2         & 0.2         \\ \cline{2-9} 
                                          & \textbf{Vh} & \cellcolor[HTML]{f9e6ff}5.1        & 2.9        & 2.8        & \cellcolor[HTML]{f9e6ff}4.2        & 1.2        & \cellcolor[HTML]{bf00ff} 96.4        & 0.3         \\ \cline{2-9} 
                                          & \textbf{Pe} & 0.8        & 0.4        & 0.0        & 1.9        & 0.4        & 0.3         & \cellcolor[HTML]{bf00ff} 98.9        \\ \hline
\end{tabular}

\begin{tablenotes}
        \item[1] Semantic classes: Building (B), Pole (P), Road (R), Undrivable Road (U), Vegetation (V), Vehicle (Vh) and Pedestrian (Pe). 
\end{tablenotes}
\end{table}

Table \ref{table:CM_single_scan} shows the normalized confusion matrix for the point classification output from the proposed pipeline. There are a few important observations from this matrix. Because the lidar can partially see through the leaves and branches of trees, the objects behind (such as \textit{building}) are sometimes mislabelled as \textit{vegetation}.
Around $7\%$ of points classified by the CNN as \textit{road} actually belong to the \textit{undrivable road} class. This is mainly due to the lack of clear boundaries between these two classes in some parts of the dataset, and the lack of these cases in the CNN training data. 
Out of all actual \textit{road} points, around $2.8\%$ are labelled as \textit{vehicle}, since in some cases the CNN classifies the vehicle shadow as part of the vehicle.

Qualitative evaluation for the projection to a single lidar scan is shown in Fig. \ref{fig:six_uncert}. In Fig. \ref{sub_es} we can observe the images of the five cameras stitched and semantically segmented. Fig. \ref{sub_p_b}-\ref{sub_p_v} depict the probability of each class in the point cloud domain in a top-down view. There is a strong correspondence between the high probability (in purple) and the actual object in the 3D world. One minor exception to this can be seen in the \textit{vegetation} class, where the existence of mislabeled points assigned a high probability of being \textit{vegetation} actually belong to a building. This occurs because of a number of instances where the camera can see the building between the branches and leaves of the vegetation, but the semantic segmentation incorrectly identifies the gaps as vegetation instead of building.

The previously generated point cloud was set as an input of the adapted Octomap framework to generate a voxelized (10 cm resolution) representation of the environment. A total of 1000 scans were used to build the map. The map registration is done by synchronizing the point clouds with the vehicle odometry. Each of the three projection strategies outlined earlier were used to generate a semantic voxelized map. These maps were then compared to a hand labelled ground truth map.

Table \ref{table:MapEvaluation} shows the evaluation metrics for the registration of the semantic point cloud. The tendency of the F1 score is to behave similarly to the single scan evaluation. The F1 score becomes larger with the addition of motion correction and occlusion handling. The recall is positively affected mostly by the motion correction process and the precision by the occlusion handling through the proposed masking technique.  

Overall, the results for the registered point cloud differ from the single scan since the former are slightly worse for semantic classes which belong to static objects, and significantly lower for dynamic classes. The recall and precision are affected by the registration process, the reduction in those metrics is due to factors mostly inherent from the sensors types.
 The vehicle used to collect the data was driven at around 16 m/s and travelled around 1.6 m per scan.  Due to the lidar's sparsity, some voxels are only initialized but not updated, relying exclusively on the very first label projected from the CNN's output especially for points located far from the sensors.
  
 The presence of dynamic objects registered by the camera and not the lidar can also affect the accuracy of the final map. The cameras capture a snapshot of the surroundings while the lidar scans the environment with rotating lasers. Each packet within the point cloud has to be synchronised (due to the motion of the platform) to compensate for differences between the point timestamp and the image timestamp. Nevertheless, when an object in the environment is moving the motion correction procedure cannot account for these changes in the corresponding image/point cloud, as both sensors perceive the environment in different ways. Certain classes such as \textit{building}, \textit{road}, \textit{vegetation} and \textit{vehicle} (that are parked) have an F1 score that is comparatively higher because in our dataset these classes are primarily static objects. The fact that most of the dynamic obstacles are in front of the class \textit{undrivable road} affects both recall and precision.

\begin{table}[t]
\centering
\caption{\small Normalized Confusion Matrix: semantic map built with our pipeline (projection + motion correction + mask)}
\label{table:CM_scan_regis}
\begin{tabular}{|l|c|c|c|c|c|c|c|c|}
\hline
\multicolumn{2}{|l|}{\multirow{2}{*}{}}                 & \multicolumn{7}{c|}{\textbf{True Label}}                                                   \\ \cline{3-9} 
\multicolumn{2}{|l|}{}                                  & \textbf{B} & \textbf{P} & \textbf{R} & \textbf{U} & \textbf{V} & \textbf{Vh} & \textbf{Pe} \\ \hline
\multirow{7}{*}{\rotatebox[origin=c]{90}{\textbf{Predicted Label}}} & \textbf{B}  &\cellcolor[HTML]{cc33ff} 76.9       & \cellcolor[HTML]{ecb3ff} 12.9       & 0.1        & \cellcolor[HTML]{f2ccff} 9.8        & \cellcolor[HTML]{f9e6ff}4.3        & 0.5         & \cellcolor[HTML]{f9e6ff}5.3         \\ \cline{2-9} 
                                          & \textbf{P}  & 0.2        & \cellcolor[HTML]{cc33ff} 68.1       & 0.1        & 1.5        & 0.2        & 0.1         & 2.6         \\ \cline{2-9} 
                                          & \textbf{R}  & 0.2        & 2.6        & \cellcolor[HTML]{cc33ff} 73.4       & \cellcolor[HTML]{f9e6ff}6.8        & 0.1        & 1.5         & 0.0         \\ \cline{2-9} 
                                          & \textbf{U}  & 0.4        & 2.8        & 2.4        & \cellcolor[HTML]{d966ff} 41.5       & 1.1        & 0.4         & 0.0         \\ \cline{2-9} 
                                          & \textbf{V}  & \cellcolor[HTML]{ecb3ff} 15.5       & \cellcolor[HTML]{ecb3ff} 12.9       & 0.4        & 3.5        & \cellcolor[HTML]{bf00ff} 92.0       & 0.1         &\cellcolor[HTML]{f9e6ff} 5.3         \\ \cline{2-9} 
                                          & \textbf{Vh} & \cellcolor[HTML]{f9e6ff}6.7        & 0.3        & \cellcolor[HTML]{e699ff} 23.4       & \cellcolor[HTML]{df80ff} 34.4       & 2.2        &\cellcolor[HTML]{bf00ff} 97.0        & \cellcolor[HTML]{df80ff}31.5        \\ \cline{2-9} 
                                          & \textbf{Pe} & 0.1        & 0.4        & 0.2        & 2.5        & 0.1        & 0.4         & \cellcolor[HTML]{d24dff}55.3        \\ \hline
\end{tabular}

\begin{tablenotes}
        \item[1] Semantic classes: Building (B), Pole (P), Road (R), Undrivable Road (U), Vegetation (V), Vehicle (Vh) and Pedestrian (Pe). 
\end{tablenotes}

\end{table}

Table \ref{table:CM_scan_regis} shows the normalized confusion matrix for the registered point cloud in the form of a semantic Octomap. In this matrix, we can see that the class \textit{building}, has 15.5$\%$ of its points mislabelled as the \textit{vegetation}. The CNN tends to classify all the pixels within trees boundaries as \textit{vegetation}, whether or not another object is visible in the background between the parts of the tree. 
The existence of obstacles that can not be detected by the lidar (like black cars, vehicle's windows, etc.) affects the precision of the class \textit{vehicle}. 
The confusion matrix indicates the percentage of points with the \textit{vehicle} label actually belong to other classes such as \textit{road}, \textit{undrivable} \textit{road} and \textit{pedestrian}.

If a projected 3D point is occluded by any of an undetected material (low reflectively) the masking technique is insufficient, keeping false positives points.

We have also calculated the percentage of labels from all the points in a single scan and voxels within the registered map. The percentages don't vary much per specific class. The class \textit{building} reduces its percentage by 0.02$\%$. From the confusion matrices, we can infer that there's a bigger confusion between the labels \textit{building} and \textit{vegetation} compared to the single scan case. Closer objects are more likely to update the same voxel since the point cloud is denser, this is the case with the class vehicle. Pedestrians, since they are dynamic are more likely to be removed from the map from future observations due to the raycasting performed by the Octomap approach.

\begin{figure}[!h]
\vspace{3mm}
\centerline{
\includegraphics[width=0.99\columnwidth]{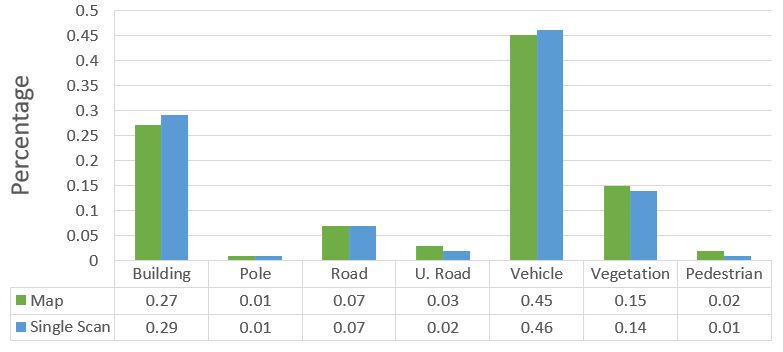}
}
\caption{\small Percentage distribution of points based on their semantic class.}
\label{fig:percent}
\end{figure}

\begin{figure}[!h]
\vspace{3mm}
\centerline{
\includegraphics[width=0.99\columnwidth]{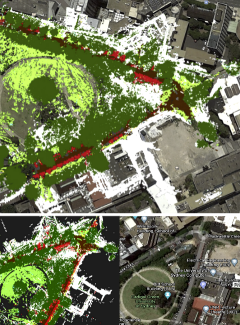}
}
\caption{\small Top-down view of the mapped area.}
\label{fig:map}
\end{figure}

As a result of the point cloud registration using the modified Octomap framework, we obtained a point cloud corresponding to the centre of each voxel. Each 3D point contains information about the location and its semantic class probability. Fig \ref{fig:map} shows the final  top-down view of the point cloud and the aerial image of the zone. We colored the points based on the semantic class with the highest probability.

\section{Conclusions and future work}
In this paper we presented a innovative pipeline that generates a probabilistic semantic octree map using various sensor modalities (camera, lidar, IMU, wheel encoders). We provide an outline of the various challenges involved in this process and present strategies to solve these problems while incorporating uncertainty at each stage.


Images captured by a vehicle were segmented by a trained CNN. The output of this network was incorporated into a new algorithm that evaluates the heuristic probabilities for each class.  These probabilities are calculated by combining the superpixel segmentation of the raw image, the network output layer and score maps. The outcome of the process is analytical probabilities for every semantic class per pixel in the image domain. 

We presented an approach to correct the lidar pointcloud for the vehicle egomotion. 
This involves applying a motion correction per lidar packet, and projecting the point cloud to the reference timestamp of the camera.
The final result is the corrected points with associated motion uncertainty information preserved for subsequent processing. 
The approach is also valid for other lidar perception applications that require probabilistic information.

From the results presented in Tables \ref{table:SingleEvaluation} and \ref{table:MapEvaluation}, the importance of applying motion correction to the lidar point cloud becomes apparent. 
The precision of the resulting point cloud projection improves significantly, as this process improves the correspondence between the each point and the image resulting in more true positives and less false positives. 
 
An efficient masking process to detect occlusions caused by the differences in the sensor mounting locations is presented in this paper. 
To accurately project the information from the image to the lidar we have to first check if each of the obstacles detected through the lidar point cloud can be observed by the camera. 
A validation process is used to determine if there is an occlusion between the camera and lidar points, in which case the points are discarded to prevent incorrect associations.
The proposed approach is designed particularly to work with lidars with low vertical resolution (such as the Velodyne VLP-16), where the occlusions could not be measured using traditional methods.

The uncertainty estimated for the semantic class within the image frame is then used to obtain a class probability distribution for each point in the 3D point cloud.
This is done using the covariance matrix describing the uncertainty in the motion correction process and the heuristic probability used to describe the labeling process. 
The uncertainty is obtained using a bivariate probability density function. 
The resulting point cloud in now extended with a set of attributes describing the probability distribution for the semantic class of each point considering the uncertainty of the labeling process and the motion correction process. 
The uncertainty of the intrinsic and extrinsic calibration parameters could also potentially be included in this process.

An octree map was built based on the vehicle odometry and the probabilistic semantic point cloud. 
Modifications to the Octomap framework allowed us to incorporate and update the semantic probabilities. 
The outcome of this work is relevant for autonomous vehicle applications as it addresses the real world challenges for generating a high level understanding of a scene.
The probabilistic nature of the resulting semantic map can be used for a risk based approach to localisation, path planning and navigation. 
The resulting 3D map has probabilistic information of occupancy and semantic labels which can be also be used to update the map when the area is revisited. 



\section*{Acknowledgment}

This work has been funded by the ACFR, the University of Sydney through the Dean of Engineering and Information Technologies PhD Scholarship (South America) and the Australian Research Council Discovery Grant DP160104081 and University of Michigan / Ford Motors Company Contract ``Next generation Vehicles".

\appendix  
\section*{lidar-camera projection} \label{lidar-camera-fnc}
The lidar-camera projection function first makes use of the generic pinhole camera-image projection equations, which states:

\begin{align}
a &=  \frac{x^{cam}}{z^{cam}} & b &=  \frac{y^{cam}}{z^{cam}} \label{eq_pc1}
\end{align}

\begin{align}
r &= \sqrt{a^{2}+b^{2}} & \theta &= \textup{atan}(r) \label{eq_pc2}
\end{align}

Since our cameras have fisheye lenses, we need apply the distortion established by the camera model to find the corresponding pixel in the image \cite{opencv}. The distortion of the lens is calculated as follows: 

\begin{equation} \label{eq_pc3}
   \theta_d = \theta(1+k_1\theta^2+k_2\theta^4+k_3\theta^6+k_4\theta^8)
\end{equation}

where $k_1$, $k_2$, $k_3$ and $k_4$ are the lens' distortion coefficients. Then we compute the distorted point coordinates as:

\begin{align}
x' &=  (\theta_d/r)a & y' &=  (\theta_d/r)b
\end{align}

The definite pixel coordinates vector \(\begin{bmatrix} u & v \end{bmatrix}^{T}\) in the image frame of a 3D lidar point can be calculated as:

\begin{align}
u &= f_x(x'+\alpha y')+c_x & v &= f_y(y')+c_y
\label{eq_pc4}
\end{align}

where $\alpha$ is the camera's skew coefficient, $[c_x,c_y]$ the principal point offset and $[f_x, f_y]$ are the focal lengths expressed in pixel units.

\section*{Bivariate probability density function}

The bivariate probability density function is defined as: 

\begin{align} \label{eq16}
f(u,v,\Sigma _{uv} )= \frac{1}{2\pi\sigma _u\sigma_v \sqrt{1-\rho ^2} }\textup{exp}\left ( -\frac{1}{2(1-\rho ^2)}p_{d} \right )
\end{align}

where $p_{d}$
\begin{align} \label{eq17}
p_{d} = \frac{(u-\mu _u)^2}{\sigma _u^2} + \frac{(v-\mu _v)^2}{\sigma _v^2} - \frac{2\rho (u-\mu _u)(v-\mu _v)}{\sigma _u^2 \sigma _v^2}
\end{align}

being $\mu _u$ and $\mu _v$ the pixel coordinate of the direct lidar-image projection, $\sigma _u$ and $\sigma _v$ the standard deviation of $u$ and $v$ and $\rho$ their correlation.


%

\ifCLASSOPTIONcaptionsoff
  \newpage
\fi



%

%

\bibliography{main}
\bibliographystyle{IEEEtran}

\begin{IEEEbiography}[{\includegraphics[width=1in,height=1.25in,clip,keepaspectratio]{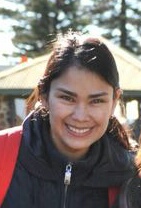}}]{Julie Stephany Berrio} received the B.S. degree in Mechatronics Engineering in 2009 from Universidad Autonoma de Occidente, Cali, Colombia, and the M.E. degree in 2012 from the Universidad del Valle, Cali, Colombia. She is currently working towards the Ph.D. degree at the University of Sydney, Sydney, Australia. Her research interest includes semantic mapping, long-term map maintenance, and point cloud processing.
\end{IEEEbiography}

\begin{IEEEbiography}[{\includegraphics[width=1in,height=1.25in,clip,keepaspectratio]{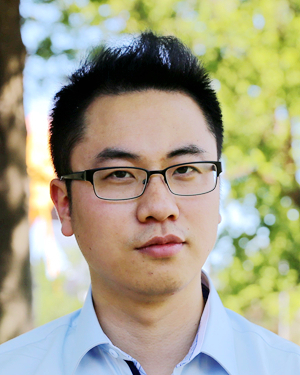}}]{Mao Shan} received the B.S. degree in electrical engineering from the Shaanxi University of Science and Technology, Xi’an, China, in 2006, and the M.S. degree in automation and manufacturing systems and Ph.D. degree from the University of Sydney, Australia, in 2009 and 2014, respectively. He is currently a Research Fellow with the Australian Centre for Field Robotics, the University of Sydney. His research interests include autonomous systems, localisation, and tracking algorithms and applications.
\end{IEEEbiography}

\begin{IEEEbiography}[{\includegraphics[width=1in,height=1.25in,clip,keepaspectratio]{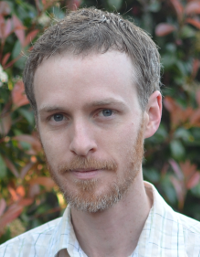}}]{Stewart Worrall} received the Ph.D. from the University of Sydney, Australia, in 2009. He is currently a Research Fellow with the Australian Centre for Field Robotics, University of Sydney. His research is focused on the study and application of technology for vehicle automation and improving safety.
\end{IEEEbiography}

\begin{IEEEbiography}[{\includegraphics[width=1in,height=1.25in,clip,keepaspectratio]{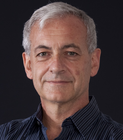}}]{Eduardo Nebot} received the BSc. degree in electrical engineering from the Universidad Nacional del Sur, Argentina, M.Sc. and Ph.D. degrees from Colorado State University, Colorado, USA. He is currently a Professor at the University of Sydney, Sydney, Australia, and the Director of the Australian Centre for Field Robotics. His main research interests are in field robotics automation and intelligent transport systems. The major impact of his fundamental research is in autonomous systems, navigation, and safety.
\end{IEEEbiography}




\end{document}